\documentclass[draftcls, onecolumn]{IEEEtran}

\usepackage{amsmath,amsfonts,amssymb}
\usepackage{multirow}
\usepackage{graphicx}
\usepackage{subfigure}
\usepackage[usenames,dvipsnames,svgnames,table]{xcolor}
\usepackage{algorithmic}
\usepackage{algorithm}
\usepackage{tablefootnote}
\usepackage{threeparttable}
\usepackage{booktabs}
\usepackage{amsthm}

\newcommand{\mbf}[1]{\mathbf{#1}}
\newcommand{\scalprod}[2]{\left\langle #1,#2 \right\rangle}

\newcommand{\red}[1]{\textcolor{red}{#1}}

\newcommand{\bb}[1]{\mathbb{#1}}

\newtheorem{theorem}{Theorem}
\newtheorem{definition}{Definition}

\renewcommand{\algorithmicrequire}{\textbf{Input:}}
\renewcommand{\algorithmicensure}{\textbf{Output:}}

\newcommand{\argmax}{\operatornamewithlimits{argmax}}

\include{notations}

\begin{document}

%\twocolumn[
%
%\aistatstitle{ Multi-task additive models with shared transfer functions}
%
%\aistatsauthor{ Anonymous Author 1 \And Anonymous Author 2 \And Anonymous Author 3 }
%
%\aistatsaddress{ Unknown Institution 1 \And Unknown Institution 2 \And Unknown Institution 3 } ]

% \title{Multi-task additive models with shared transfer functions}
\title{Multi-task additive models with shared transfer functions based on dictionary learning}

\author{Alhussein~Fawzi,
        Mathieu Sinn,
        and Pascal Frossard
\thanks{A. Fawzi and P. Frossard are with Ecole Polytechnique Federale de Lausanne (EPFL), Signal Processing Laboratory (LTS4), Lausanne, Switzerland (e-mail: alhussein.fawzi@epfl.ch; pascal.frossard@epfl.ch).}
\thanks{M. Sinn is with IBM Research, Dublin, Ireland.}}% <-this % stops a space

%\title{\Large Multi-task additive models with shared transfer functions}
%\author{Alhussein Fawzi\thanks{Signal Processing Laboratory (LTS4), EPFL -- Lausanne, Switzerland. \{alhussein.fawzi, pascal.frossard\}@epfl.ch} \thanks{This work was partly done during the author's internship at IBM Research Ireland.}  \\
%\and 
%Mathieu Sinn\thanks{IBM Research -- Ireland. mathsinn@ie.ibm.com}
%\and
%Pascal Frossard\samethanks[1]
%}
%\date{}

\maketitle

\begin{abstract} % \small\baselineskip=10pt
Additive models form a widely popular class of regression models which represent the relation between covariates and response variables as the sum of low-dimensional \textit{transfer functions}. Besides flexibility and accuracy, a key benefit of these models is their \textit{interpretability}: the transfer functions provide visual means for inspecting the models and identifying domain-specific relations between inputs and outputs. However, in large-scale problems involving the prediction of many related tasks, learning independently additive models results in a loss of model interpretability, and can cause overfitting when training data is scarce. % Moreover, when training data is scarce, an independent learning of the different tasks can lead to overfitting.
% and performance, when training data is scarce. 
We introduce a novel multi-task learning approach which provides a corpus of accurate and interpretable additive models for a large number of related forecasting tasks. Our key idea is to \textit{share transfer functions across models} in order to reduce the model complexity and ease the exploration of the corpus. We establish a connection with sparse dictionary learning and propose a new efficient fitting algorithm which alternates between sparse coding and transfer function updates. The former step is solved via an extension of Orthogonal Matching Pursuit, whose properties are analyzed using a novel recovery condition which extends existing results in the literature. The latter step is addressed using a traditional dictionary update rule. Experiments on real-world data demonstrate that our approach compares favorably to baseline methods while yielding an interpretable corpus of models, revealing structure among the individual tasks and being more robust when training data is scarce. Our framework therefore extends the well-known benefits of additive models to common regression settings possibly involving thousands of tasks.
% which generalizes state-of-the-art results in the field. 
%
%We develop an efficient fitting algorithm for the novel model and show that 
%
%multi-task additive model with shared transfer functions, that allows to 
%
% a poor prediction accuracy when the number of training examples is limited.
%does not give an interpretable 
%
%
%We consider the problem of forecasting many related tasks using additive model regression.  
%
%We consider in this paper the problem of  multi-task learning for additive models with a shared underlying representation.
%% that accurately model the data. We consider in this paper 
%
%that extract key patterns and trends
%is to make sense of the data by extracting key patterns and trends.
%
%With the huge volumes of data produced by various sensors, one of the major challenges
%is to make sense of the data by extracting key patterns and trends.
%\red{With the advent of big data, new processing tools needed etc... We have many signals, and we want to find a small number of models that fit them well all. We frame the problem as a dictionary learning problem, and leverage the recent advances in this topic to propose an efficient solution. We illustrate our approach on electricity load forecasting. We propose an algorithm. Experiments on many signals. We also illustrate our technique can be used for intra-signal clustering. We use a modified version of OMP, and the theoretical guarantees are useful independently of the other ...}
\end{abstract}
\begin{IEEEkeywords}
Additive models, nonparametric regression, dictionary learning, sparse representations, multi-task learning.
\end{IEEEkeywords}

\section{Introduction}\label{sec:introduction}
% In regression problems, the goal is to estimate the relashionships between variables. Linear regression models, which models the response variable as a linear function of the covariates, has been the focus of much research. 
% an adequate and interpretable description of how the inputs affect the output.
% Having transfer functions that are shared across the tasks allows to model the different effects of the covariates on the response variables 
 % is common to a large number of tasks.
%The focus of regression is to model the relationship between a response variable and a set of covariates. The linear model, %which represents the response variable as a linear function of the covariates, gives an interpretable description of how the inputs %affect the output. However, many regression problems cannot be modeled linearly and require more complex nonlinear models. 
Additive models are a widely popular class of nonparametric regression models which have been extensively studied theoretically and successfully applied to a wide range of practical problems in signal processing and machine learning \cite{hastie2009elements,wood2006generalized,hastie1990generalized}. The key ingredient of additive models are \textit{transfer functions} that explain the effect of covariates on the response variable in an additive manner. % ; a common approach is to use spline basis expansions for their representation, thereby allowing for the modeling of nonlinear effects. 
Besides being flexible (e.g., allowing for the modeling of nonlinear effects for both continuous and categorical covariates) and yielding good predictive performance, an important selling point of additive models is their interpretability. In particular, the transfer functions provide intuitive visual means for application experts to understand the models and explore the relationship between input and output signals of the system under study.

In many real-world data modeling settings, one faces the problem of forecasting a large number (e.g., several thousands) of related tasks. % An example that we are studying in this paper is predicting the electricity demand for a large number of individual customers equipped with smart meters. 
In this case, learning additive models independently for each task has several disadvantages. Firstly, the number of models would be too large for a domain expert to visually inspect all the transfer functions, hence - in essence - the corpus of models loses its interpretability from a human point of view. Secondly, independently learning the models ignores \textit{structure} and \textit{commonality} among the tasks.
%, e.g., classes of customers that exhibit similar demand patterns; 
Thirdly, when training data is scarce, learning the models independently is prone to overfitting the data.

To overcome these challenges, we introduce a novel \textit{multi-task learning} framework for additive models. Intuitively, the key idea is to \textit{share transfer functions across tasks} that exhibit commonality in their relationships between input and output variables. More specifically, each individual task is modeled as a weighted sum of transfer functions chosen from a candidate set which is common to all tasks, and the cardinality of which is small relative to the total number of tasks. 
\iffalse
In many problems of interest, the effect of covariates on response variables obeys the same law for a large number of tasks. For example, customers from the same neighborhood that use the same type of heating (e.g., gas or electric) are share similar consumption patterns with respect to temperature, and are therefore likely to share the ``temperature'' transfer function.
% \red{Probably add sentence here on segmentation} 
We propose in this paper a novel multi-task learning framework for additive models with a \textit{shared underlying representation}. Specifically, the response variable of each task is modeled as an additive combination of univariate transfer functions, where transfer functions are further constrained to belong to sets of candidate functions. These sets -- one per covariate -- are \textit{shared} across all tasks, and contain the different transfer functions that represent the potential effects of covariates on the response variables. To capture the specificity of each task, we further allow transfer functions to be scaled by \textit{task-dependent} positive weights. 
\fi
Our algorithm for solving the multi-task additive model learning problem uses an intrinsic connection with \textit{sparse dictionary learning} \cite{aharon2006svd, tosic2011dictionary, kreutz2003dictionary}. More specifically, we reformulate the fitting problem as a special form of dictionary learning with additional constraints; leveraging recent advances in the field, we propose a novel fitting approach that alternates between updates of the transfer functions and the weights that scale these functions. We introduce a novel algorithm for updating the coefficients that scale the transfer functions, called Block Constrained Orthogonal Matching Pursuit (BC-OMP), which extends conventional Orthogonal Matching Pursuit \cite{pati1993orthogonal, Mallat93}. Furthermore, we derive novel coherence conditions for the accurate recovery of the optimal solution which are interesting in their own right as they extend existing theory. Transfer functions, which correspond to dictionary elements in our dictionary learning analogy, are updated using a traditional dictionary step update \cite{engan1999method}.
% This algorithm is shown to recover accurately the optimal solution, under some conditions on the transfer functions.

In the experimental part of our paper, we apply the proposed algorithm to synthetic and real-world electricity demand data. Synthetic results show that the proposed approach accurately learns transfer functions from noisy data. In addition, the proposed method is shown to outperform baseline linear and non-linear regression methods in terms of prediction accuracy. In a second experiment, we use a dataset of 4,066 smart meter time series data from Ireland, and show that our approach yields predictive performance comparable to baseline methods while only using a small number of candidate functions; interestingly, the discovered commonality of tasks corresponds to classes of residential and different types of enterprise customers. When using only a small fraction of the training data, our approach yields more robust results than independent learning and hence inherits the benefits of traditional multi-task learning. In a final experiment, we apply our multi-task learning algorithm to a \textit{single-task} problem to improve the prediction accuracy of traditional additive model learning, while maintaining the number of learned transfer functions small. % We finally present an original application of our multi-task learning framework, where the algorithm is applied to a \textit{single} task to improve the prediction accuracy of traditional additive model learning, while maintaining the number of learned transfer functions small.
Over the past decade, many works have shown the benefits of multi-task learning over independently learning the tasks \cite{caruana1997multitask, bakker2003task, Thrun_1997_622}, and different approaches to multi-task learning were considered.
% In the context of multi-task learning \cite{caruana1997multitask}, many approaches have been considered previously. 
% The problem of multi-task learning \cite{} has recently received significant attention from researchers in the field, and led to different approaches. 
In \cite{evgeniou2005learning}, the authors impose the linear weight vectors of different tasks to be close to each other. % of the learned linear predictors for each task to be close to each other.
% We highlight some of the related works in the multi-task learning. In the context of multi-task learning The problem of multi-task learning has received a lot of attention recently 
% Previous work in this direction is \cite{evgeniou2005learning} who impose the weight vectors of the learned linear predictors to be close to each other.
%, in the context of linear models, impose the model parameters to be close to each other with respect to a given norm.
% obozinski2006multi
The work in \cite{evgeniou2007multi} constrains the weight vectors to live in a low-dimensional subspace. Still in the context of linear models, the authors of \cite{jacob2009clustered} assume that the tasks are clustered into groups, and that tasks within a group have similar weight vectors.
% the enforce the predictors for different tasks to use similar features from a (potentially huge) candidate set.
In the context of additive models, \cite{liu2009nonparametric} proposes new families of nonparametric models that enforce the selected covariates to be the same across tasks. This setting is particularly relevant for regression tasks involving a large number of covariates $p$, and the algorithm in \cite{liu2009nonparametric} extracts a common set of covariates for the tasks. Our work significantly differs from \cite{liu2009nonparametric} in several aspects. While \cite{liu2009nonparametric} enforces a common set of covariates across tasks, the transfer functions are different. In other words, their approach only leverages commonality with respect to \textit{which} covariates affect the dependent variable, but not \textit{how} they affect it, leading to a number of transfer functions that is still too large for inspection by human experts. By imposing a common set of candidate transfer functions across tasks, we limit the number of transfer functions, and obtain interpretable models even for problems involving thousands of tasks. Moreover, unlike \cite{liu2009nonparametric}, we consider a setting where all covariates are relevant for the task at hand. Hence, in this paper, we are typically interested in problems involving a small number of input covariates $p$, and a very large number of tasks $N$. Our approach shows that, by learning a number of transfer functions that is much smaller than $pN$, it is possible to achieve comparable or better performance than models involving a much larger number of parameters.
% assumes that all $p$ features are relevant for the task at hand, and does not assumes extract a sparse set of features out of the original covariates.
% only leverages commonality with respect to \textit{which} covariates affect the dependent variable, but not \textit{how} they affect it.
% considers additive models where the selected covariates are enforced to be the same across tasks, but the transfer functions can be different. While this results in an improved predictive performance for problems with a large number of covariates, the number of transfer functions may still be too large for inspection by human experts; moreover, this approach only leverages commonality with respect to \textit{which} covariates affect the dependent variable, but not \textit{how} they affect it. \red{add sentence that outlines the benefits of our method}

The paper is structured as follow: Sec.~\ref{sec:additive_model_review} introduces notation and provides a review of additive models.
In Sec.~\ref{sec:multitask model formulation} we formulate the multi-task additive model learning problem and establish the connection with sparse dictionary learning.
The algorithm for solving the multi-task problem is explained in Sec.~\ref{sec:learning algorithm}, and the theoretical analysis of recovery conditions is provided in Sec.~\ref{sec:recovery_cond_COMP}. In Sec.~\ref{sec:experiments} we describe our experiments on synthetic and real-world data; conclusions and an outlook on future research are given in Sec.~\ref{sec:conclusions}.

\section{Preliminaries}

% \red{Add bias in additive model!!}

% \red{Say that we have identifiability problem. Therefore, introduce the centered basis functions, say that the original problem is equivalent to the new unconstrained one.}

\label{sec:additive_model_review}

% \noindent \textbf{Notations:} 

\subsection{Notations} We use \textbf{boldface} notation for vectors and matrices. Moreover, we use $[n]$ to refer to the set $\{1, \dots, n\}$. % Given a set $\{ \lambda_{lj} \}_{l \in [a], j \in [b]}$ of real numbers, we define, for any fixed $l$, $\boldsymbol\lambda_l$ as the vector $[\lambda_{l1}, \dots, \lambda_{lb}]^T$ and, for any fixed $j$, $\boldsymbol\lambda_{,j}$ as $[\lambda_{1j}, \dots, \lambda_{aj}]^T$. 
% Besides, we use the notation $\triangleq$ when we define a new quantity. 
Given a vector $\mbf{z}$, we denote by $\| \mathbf{z} \|_0$ the $\ell_0$ ``norm'', that counts the number of nonzero elements in $\mbf{z}$. Also, we denote by $\otimes$ the Kronecker product operation. If $\mbf{Z} \in \mathbb{R}^{n_1 \times n_2}$ is a matrix, $\text{vec} (\mbf{Z}) \in \mathbb{R}^{n_1 n_2}$ denotes the vectorization of $\mbf{Z}$, obtained by stacking the columns of $\mbf{Z}$, and $\mbf{Z}^\dagger$ denotes the Moore-Penrose pseudo-inverse. Moreover, we use the notation $\mbf{z} \in \mathbf{Z}$ to denote that $\mbf{z}$ is one of the columns of $\mathbf{Z}$. Finally, $\mathbf{Z} \geq \mathbf{0}$ denotes the entry-wise non-negativity constraint.

\subsection{Additive models review}
\label{sec:additive_model_review_subpart}

\begin{figure}[t]
\centering
\includegraphics[width=0.3\textwidth]{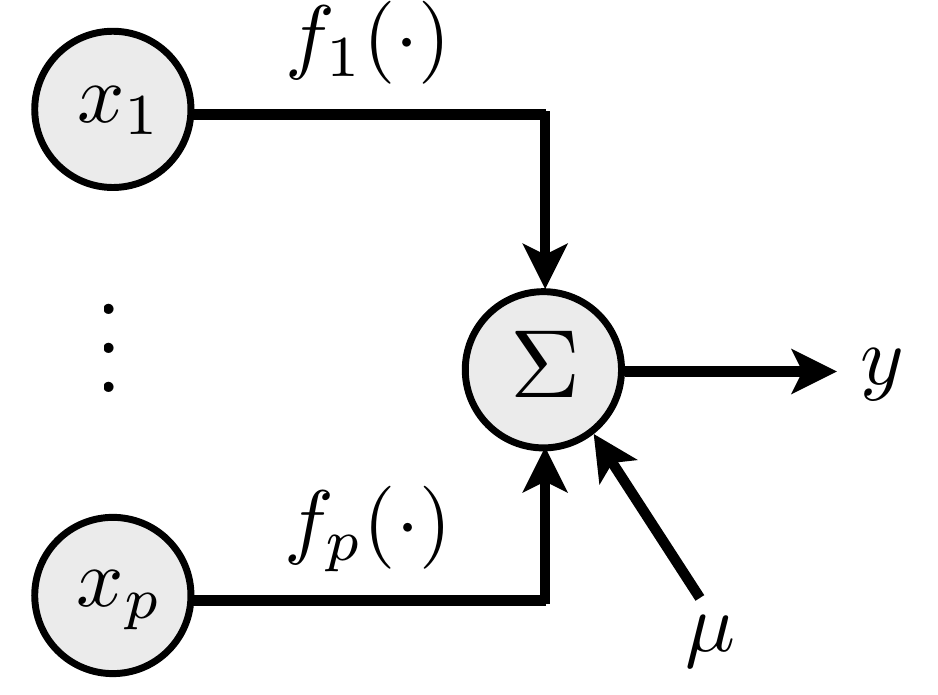}
\caption{\label{fig:diagram_am} Additive model diagram}
\end{figure}

We first briefly review additive models. Let $\{ x_{ij} , i  \in [n], j \in [p] \}$ and $\{ y_i, i \in [n] \}$ denote respectively the observed covariates and response variable. Here, $n$ is the number of observations and $p$ the number of covariates. Additive models have the form:
\begin{align*}
y_i = \mu + \sum_{j=1}^p f_j(x_{ij}) + \epsilon_i,
\end{align*}
where $\mu$ is the intercept and $\epsilon_i$ is assumed to be a white noise process. The \textit{transfer functions} $f_j$ represent the effect of a covariate on the response variable. The additive model is illustrated in Fig. \ref{fig:diagram_am}. To ensure unique identification of the $f_j$'s, we assume that transfer functions are centered: $\sum_{i=1}^n f_j(x_{ij}) = 0$ for all $j \in [p]$.
Nonlinear transfer functions of continuous covariates are commonly modeled as smoothing splines \cite{wood2006generalized, hastie2009elements}, i.e.,
% \footnote{For discrete covariates, one can simply use a decomposition in the canonical basis.}
\begin{align}
\label{eq:spline_eq}
f_j(z) = \sum_{t=1}^{T_j} \beta_{jt} \phi_{jt} (z),
\end{align}
where $\beta_{jt}$ denotes the spline coefficients, $\phi_{jt}$ the B-spline basis functions, and $T_j$ the number of basis splines. % Note that $T_m$ is intimately related to the number of knot points. % $q_m$ through the expression $T_m = q_m - d - 1$, where $d$ is the degree of the splines. 
Using this representation, estimating the transfer functions therefore amounts to the estimation of the spline coefficients $\beta_{jt}$ and the intercept $\mu$. We consider the following fitting problem with centering constraints: 
\begin{align*}
& \min_{\mu,\{\beta_{jt}\}} \sum_{i=1}^n \left( y_i - \mu - \sum_{j=1}^p \sum_{t=1}^{T_j} \beta_{jt} \phi_{jt} (x_{ij}) \right)^2 \\ & \text{ subject to } \sum_{i=1}^n \sum_{t=1}^{T_j} \beta_{jt} \phi_{jt} (x_{ij}) = 0 \text{ for all } j \in [p].
\end{align*}
One can convert the above problem to an unconstrained optimization problem by centering the response and basis functions. Specifically, let $\bar{\phi}_{jt} = \frac{1}{n} \sum_{i=1}^n \phi_{jt} (x_{ij})$, $\mbf{s}_j(z) = [\phi_{j1} (z) - \bar{\phi}_{j1}, \dots, \phi_{j T_j} (z) - \bar{\phi}_{j T_j}]^T$ and
\begin{align}
\label{eq:design_matrix_nonCentered}
\mbf{S} = \begin{bmatrix} 
\mbf{s}_1 (x_{11})^T & \dots & \mbf{s}_p (x_{1p})^T \\
& \vdots & \\
\mbf{s}_1 (x_{n1})^T & \dots & \mbf{s}_p (x_{np})^T
\end{bmatrix}.
\end{align}
We define the vectorized spline coefficients $\boldsymbol\beta = \begin{bmatrix} \boldsymbol\beta_1^T & \dots & \boldsymbol\beta_p^T \end{bmatrix}^T$ with $\boldsymbol\beta_j = \begin{bmatrix} \beta_{j1} & \dots & \beta_{jT_j} \end{bmatrix}^T$.
The above constrained fitting problem is then equivalent to the following unconstrained least squares problem:
\begin{align*}
\min_{\boldsymbol\beta} \| \mbf{y} - \mbf{S} \boldsymbol\beta \|_2^2,
\end{align*}
where $\mbf{y}$ denotes the centered response variables $\mbf{y} = [y_1 - \bar{y}, \dots, y_n - \bar{y}]^T$, with $\bar{y} = 1/n \sum_{i=1}^n y_i$. In order to avoid overfitting, a quadratic penalizer is commonly added, leading to the problem:
\iffalse
simple linear regression can easily lead to overfitting. A common way to overcome this problem is to use a quadratic penalizer in the objective function: %, a regularizer is commonly added to the above fitting objective function: A quadratic regularizer is therefore commonly added to the fitting term, which leads to
\fi
\begin{align*}
\min_{\boldsymbol\beta} \| \mbf{y} - \mbf{S} \boldsymbol\beta \|_2^2 + \boldsymbol\beta^T \boldsymbol\Sigma \boldsymbol\beta,
\end{align*}
with a regularization matrix $\boldsymbol\Sigma$. The penalized minimization problem has the closed form solution:
\begin{align*}
\hat{\boldsymbol\beta} = (\mbf{S}^T \mbf{S} + \boldsymbol\Sigma)^{-1} \mbf{S}^T \mbf{y},
\end{align*}
provided that $\mbf{S}^T \mbf{S} + \boldsymbol\Sigma$ is non-singular.
% can then be solved in closed form using a standard linear least squares solver.

% is obtained by solving the following penalized least-squares problem:
%\begin{align*}
%\hat{\boldsymbol\beta} = \argmin_{\boldsymbol\beta} \| \mbf{y} - \mbf{S} \boldsymbol\beta \|_2^2 + \boldsymbol\beta^T \boldsymbol\Sigma \boldsymbol\beta,
%\end{align*}
% where $\boldsymbol\Sigma$ is a regularization matrix. The above minimization problem can then be easily solved using a standard linear least squares solver.
% \red{Put here the learning problem of additive models, and how it is solved}.

\section{Multi-task additive model with shared transfer functions}\label{sec:multitask model formulation}

% We now assume a $N$-task regression problem, where $\{ \mathbf{x}_j , j  \in [n] \}$ and $\{ \mathbf{z}_j^{(i)} , j  \in [n], i \in [N] \}$  denote respectively \textit{common} and \textit{task-specific} covariates, and $\{ y_j^{(i)}, j  \in [n], i \in [N] \}$ denote the response variables. The superscript notation $(i)$ is used to denote the task index. While the observations $x_{jm}$ are common to all the tasks, $z_{jm}^{(i)}$ differ with the task, and allow to model effects that are specific to every task. Each task may depend in a different way on the covariates, leading to one response variable per task $\mbf{y}^{(i)}$. We stack the response variables of the different tasks in a global matrix $\mbf{Y} = [\mbf{y}^{(1)} | \dots | \mbf{y}^{(N)}] \in \mathbb{R}^{n \times N}$.

\begin{figure}[t]
\centering
\includegraphics[width=0.55\textwidth]{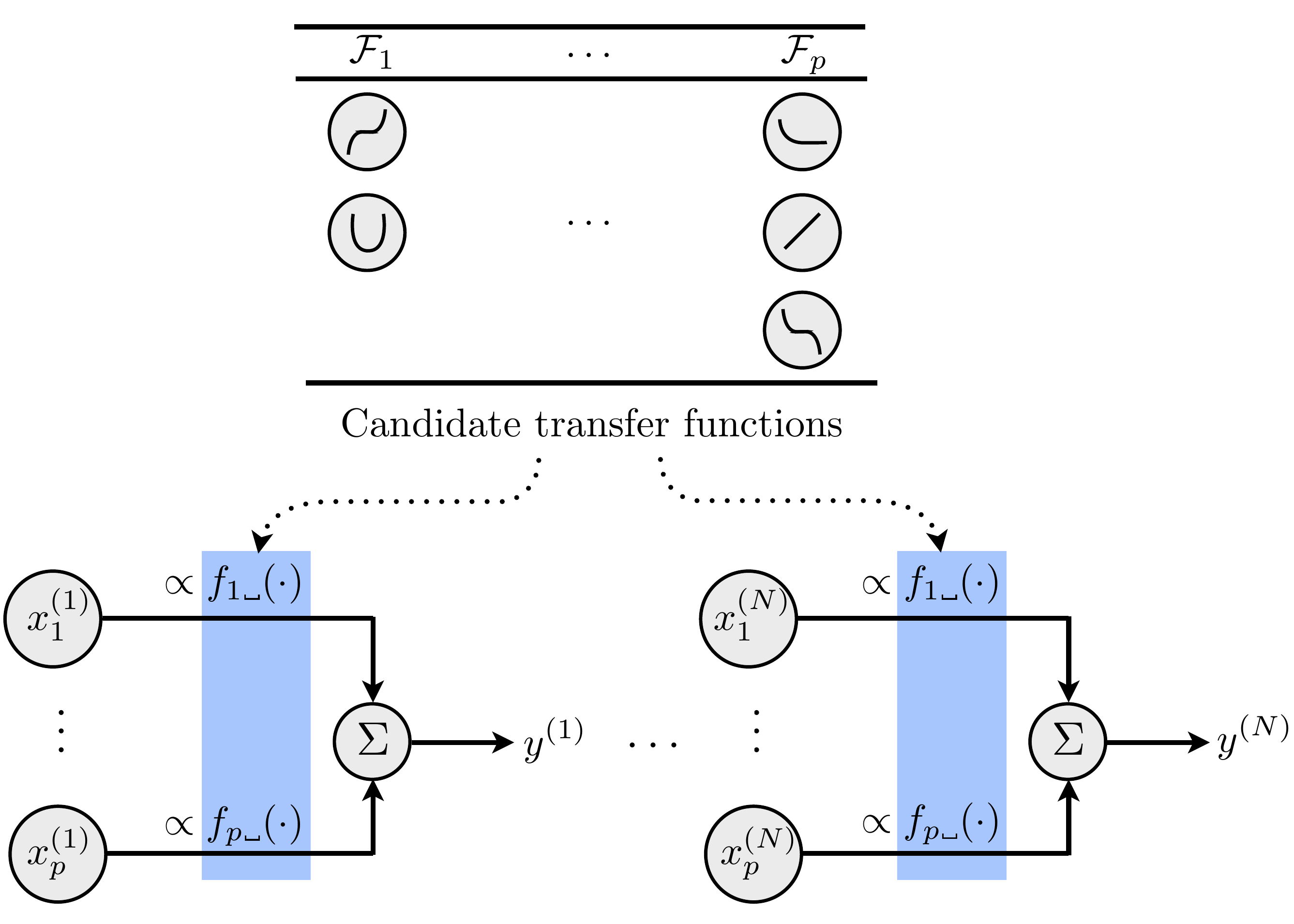}
\caption{\label{fig:diagram_multitask_am} Multi-task additive model diagram. $\mathcal{F}_j$ denotes the set of $L_j$ candidate transfer functions that model the effects of covariate $j$. The sets $\mathcal{F}_j, 1 \leq j \leq p$ are common to all tasks. Each task is modeled as a linear combination of transfer functions chosen from the sets $\mathcal{F}_1, \dots, \mathcal{F}_p$. In this diagram, the symbol \textvisiblespace{ } denotes an arbitrary index in $[L_j]$.}
\end{figure}

We now introduce our new multi-task additive model with shared transfer functions.
We assume a $N$-task regression problem where $\{ x_{ij}^{(m)} , i  \in [n], j \in [p], m \in [N] \}$ are the covariates,
and $\{ y_i^{(m)}, i  \in [n], m \in [N] \}$ denotes the response variables, where the superscript $(m)$ is the task index. % In this paper, we restrict ourselves to the case where the covariates are common to all tasks; the case where covariates are task-dependent is left open for future work.
%Since this is not central to the paper, we postpone this discussion to a future occasion. 
We further assume without loss of generality that the response variables have zero mean. 
% It is assumed here that the covariates are \textit{common} to all the tasks. 
Our multi-task model is given as follows:
% Each task depends however in a different way on the covariates, leading to one response variable per task. We stack the response variables of the different tasks in a global matrix $\mbf{Y} = [\mbf{y}^{(1)} | \dots | \mbf{y}^{(N)}] \in \mathbb{R}^{n \times N}$. 
%  Likewise, the covariate matrices $\mbf{X^{(i)}} \in \mathbb{R}^{n \times p}, i \in [N]$ stack the covariate vectors of different tasks. 
% Besides the computational burden of learning a model for each task, using independent additive models for each task results in $p N$ transfer functions, that can very difficult to interpret. Moreover, in the regime where the number of training examples is small, learning independent models might lead to overfitting.
% When the number of signals $N$ is large, using independent additive models for each signal $i \in [N]$ results in $p N$ transfer functions, that are difficult to interpret. \red{One can mention at this point the other drawbacks that I had put in the presentation also} 
% We propose here a model that allows having \textit{shared transfer functions} across the different tasks. 
% Let $L_j$ moreover denote the specified number of candidate transfer functions for covariate $j$. Then, the model is given as follows:
\begin{align}\label{eq:model_shared}
& \hspace{-63mm} y_i^{(m)} = \sum_{j=1}^p \sum_{l=1}^{L_j} \lambda_{jl}^{(m)} f_{jl} \left(x_{ij}^{(m)}\right) + \epsilon_i^{(m)}\\ 
\hspace{-1mm}  \text{ with } \| \boldsymbol\lambda_{j}^{(m)} \|_0 \leq 1 \text{ and } \boldsymbol\lambda_{j}^{(m)} \geq \mbf{0} \text{ for all } j \in [p], m \in [N]. \nonumber
\end{align}
% TO DO FOR THE JOURNAL VERSION
%\red{NOTE: We can actually augment the model so that there are some covariates that are specific to each task (i.e., a part that is specific, and a part that is common to all the tasks). The model is then given as follows:
%\begin{align}
%\label{eq:model_shared}
%& y_j^{(i)} = \sum_{m=1}^p \sum_{l = 1}^{L_m} \lambda_{lm}^{(i)} f_{lm} (x_{jm}) + \sum_{m=1}^q g^{(i)}_m (z_{jm}^{(i)})  + \epsilon_j^{(i)}, \\
%& \text{ with } \| \boldsymbol\lambda_{,m}^{(i)} \|_0 \leq 1 \text{ and } \boldsymbol\lambda_{,m}^{(i)} \geq 0 \text{ for all } m \in [p], i \in [N]. \nonumber
%\end{align}
%The algorithm can easily be adapted to take into account that (more sophisticated) model.
%}
% END TO DO FOR THE JOURNAL VERSION
% Our regression model is the sum of two distinct components. The first component consists of an additive model with shared transfer functions across the tasks. 
Note that, in our new model, the response variables are \textit{weighted} linear combinations of $p$ transfer functions, each of which is selected from the set $\mathcal{F}_j \triangleq \{ f_{jl}, l \in [L_j] \}$ which contains 
$L_j$ candidate transfer functions that model the effects of the covariate $j$. % The $\ell_0$ norm constraint constrains at most one transfer function in each set $\mathcal{F}_m$ to be active for each task.
The $\ell_0$ norm constraint on the weights $\boldsymbol\lambda_{j}^{(m)}$ prevents two transfer functions from the set $\mathcal{F}_j$ to be active for the same task. Hence, only one transfer function captures the effect of a covariate in a response variable. This constraint is crucial, as it disallows the creation of ``new'' transfer functions from the candidate ones by linearly combining them. 
% The weights $\lambda_{lm}^{(i)}$ allow to scale the transfer function adequately for every task. 
While the transfer functions $f_{jl}$ are \textit{common} to all the tasks, the non-negative weights $\lambda_{jl}^{(m)}$ are \textit{task-specific} and permit to scale the transfer functions specifically for each task. 
This offers extra flexibility as a wide range of tasks can be modeled using the model in Eq.~(\ref{eq:model_shared}) while keeping the number of (standardized) candidate transfer functions small.
As we will see in Sec.~\ref{sec:experiments}, the non-negativity constraint in Eq.~(\ref{eq:model_shared}) facilitates the interpretation of the activation of the same transfer functions across different tasks as commonality; without this constraint, the same transfer functions could represent exactly opposite effects, e.g., higher temperatures leading to higher electricity demand for one task, and leading to lower demand for another one. Our multi-task model is illustrated in Fig. \ref{fig:diagram_multitask_am}.

% NOTE: for example: tasks with response variables of different orders of magnitude, or tasks that do not have the same %importance of transfer functions.
%
%The non-negativity assumption guarantees that, by multiplying the weights with the transfer functions, the monotonicity of the %transfer functions are preserved. This constraint is key for interpretability, as the change of monoticity of a transfer function %implies a change in its interpretation. The weights in our model essentially allow transfer functions to be scale-invariant. 

Similarly to what is done with single-task additive models (Sec. \ref{sec:additive_model_review_subpart}), we model transfer functions using smoothing splines. Specifically, we write:
\begin{align}
\label{eq:spline_tfs}
f_{jl} (z) = \mbf{s}_j (z)^T \boldsymbol\beta_{jl},
\end{align}
where $\mbf{s}_j$ and $\boldsymbol\beta_{jl}$ denote the centered spline basis functions and coefficients, respectively. Using this representation, we rewrite the model in Eq.~(\ref{eq:model_shared}) in the following vector form:
\begin{align*}
\label{eq:model_shared_matrixForm}
\forall m \in [N], \quad \mbf{y}^{(m)} = \sum_{j=1}^p \mbf{S}_j^{(m)} \mbf{B}_j \boldsymbol\lambda_{j}^{(m)} + \boldsymbol\epsilon^{(m)},
\end{align*}
%
%\begin{align}
%\label{eq:model_shared_matrixForm}
%& \mathbf{Y} = \sum_{j=1}^p \mathbf{S}_j^{(m)} \mathbf{B}_j \boldsymbol\Lambda_j + \mathbf{E} \\
%& \hspace{-5mm} \text{with } \| \boldsymbol\lambda_{j}^{(m)} \|_0 \leq 1 \text{ and } \boldsymbol\Lambda_j \geq \mbf{0} \text{ for all } j,m, \nonumber
%\end{align}
where $\mathbf{S}_j^{(m)} = \begin{bmatrix} \mbf{s}_j\left(x_{1j}^{(m)}\right) & \dots & \mbf{s}_j\left(x_{nj}^{(m)}\right) \end{bmatrix}^T$, $\mathbf{B}_j = \begin{bmatrix} \boldsymbol\beta_{j1} & \dots & \boldsymbol\beta_{jL_j} \end{bmatrix}$, % $\boldsymbol\Lambda_j = \begin{bmatrix} \boldsymbol\lambda_{j}^{(1)} & \dots & \boldsymbol\lambda_{j}^{(N)} \end{bmatrix}$, $\mbf{Y} = \begin{bmatrix} \mbf{y}^{(1)} & \dots & \mbf{y}^{(N)} \end{bmatrix}$
 and $\boldsymbol\epsilon^{(m)}$ is a Gaussian iid random vector with zero mean. The model fitting then consists in finding admissible $\{ \mathbf{B}_j \}_{j=1}^p$ and $\{ \boldsymbol\lambda_j^{(m)} \}_{j \in [p], m \in [N]}$ that minimize the sum of squared residuals, while avoiding overfitting. We therefore write the problem as follows:
\begin{align*}
\text{(P): } \min_{ \substack{ \mbf{B}_j, \boldsymbol\lambda_{j}^{(m)} } }  \sum_{m=1}^N \left\| \mbf{y}^{(m)} - \sum_{j=1}^p \mathbf{S}_j^{(m)} \mathbf{B}_j \boldsymbol\lambda_{j}^{(m)} \right\|_2^2 + \Omega( \{ \mathbf{B}_j \}_{j=1}^p ), \\
\text{ subject to }  \| \boldsymbol\lambda_{j}^{(m)} \|_0 \leq 1 \text{ and } \boldsymbol\lambda_j^{(m)} \geq \mbf{0} \text{ for all } j,m,
\end{align*}
%\begin{align*}
%\text{(P): } \quad \min_{ \{ \mbf{B}_j, \boldsymbol\Lambda_j \}_{j=1}^p}  \left\| \mbf{Y} - \sum_{j=1}^p \mathbf{S}_j^{(m)} \mathbf{B}_j \boldsymbol\Lambda_j \right\|_F^2 + \Omega( \{ \mathbf{B}_j \}_{j=1}^p ), \\
%\text{ subject to }  \| \boldsymbol\lambda_{j}^{(m)} \|_0 \leq 1 \text{ and } \boldsymbol\Lambda_j \geq \mbf{0} \text{ for all } j,m,
%\end{align*}
where $\Omega$ is a regularization term that prevents model overfitting. Note that, unlike traditional additive model learning, the above problem has two types of unknowns, that is, weights and transfer functions.  In this paper, we use the following regularization function % In our experiments, we use the regularization function
\begin{eqnarray}\label{eq:regularization}
\Omega( \{ \mathbf{B}_j \}_{j=1}^p ) = \mathbf{b}^T \boldsymbol\Sigma \mathbf{b},
\end{eqnarray}
with the regularization matrix $\boldsymbol\Sigma = \nu \mathbf{I}$ and $\nu > 0$, and $\mathbf{b}$ is the vector formed by concatenating $\text{vec}(\mathbf{B}_j), 1 \leq j \leq p$. Note that this regularizer penalizes large coefficients of smoothing splines with the strength of this effect tuned by $\nu$.

The fitting problem (P) is inherently related to \textit{sparse dictionary learning} \cite{tosic2011dictionary} where the goal is to find the dictionary $\mathbf{D}$ and sparse codes $\mathbf{C}$ that minimize
\begin{align*}
\| \mbf{Y} - \mbf{D C} \|_F^2 \text{ subject to } \| \mbf{c} \|_0 \leq p \text{ for all }  \mbf{c}\in\mbf{C}, 
\end{align*}
with $\mathbf{Y} = [\mathbf{y}^{(1)} \dots \mathbf{y}^{(N)}] \in \bb{R}^{n \times N}$, and $p$ is the desired level of sparsity. To simplify the exposition of the analogy, let us consider the multi-response scenario where covariates are equal across tasks ($x_{ij}^{(m)} = x_{ij}$ for all $m$). In this case, we have $\mathbf{S}_j^{(m)} = \mathbf{S}_j$ for all $m \in [N|$. We define the subdictionaries (or \textit{blocks}) $\mathbf{D}_j \triangleq \mathbf{S}_j \mathbf{B}_j$ and the global dictionary $\mathbf{D} \triangleq \begin{bmatrix} \mathbf{D}_1 & \dots & \mathbf{D}_p \end{bmatrix}$. The problem (P) can be rewritten as follows:
%\begin{align*}
%& \left\|  \mathbf{Y} - \mbf{D} \begin{bmatrix} \boldsymbol\Lambda_1 \\ \vdots \\ \boldsymbol\Lambda_p \end{bmatrix} \right\|_F^2 + \Omega( \{ \mathbf{B}_j \}_{j=1}^p ) \\
%																					& \text{subject to }  \| \boldsymbol\lambda_{j}^{(m)} \|_0 \leq 1 \text{ and } \boldsymbol\Lambda_j \geq \mbf{0} \text{ for all } j,m.
%\end{align*}
\begin{align*}
& \left\|  \mathbf{Y} - \mbf{D} \boldsymbol\Lambda \right\|_F^2 + \Omega( \{ \mathbf{B}_j \}_{j=1}^p ) \\
																					& \text{subject to }  \| \boldsymbol\lambda_{j}^{(m)} \|_0 \leq 1 \text{ for all } j,m \text{ and } \boldsymbol\Lambda \geq \mbf{0},
\end{align*}
with 
\begin{align*}
\boldsymbol\Lambda = \begin{bmatrix} 
\boldsymbol\lambda_1^{(1)} & \dots & \boldsymbol\lambda_1^{(N)} \\
\vdots & \vdots & \vdots \\
\boldsymbol\lambda_p^{(1)} & \dots & \boldsymbol\lambda_p^{(N)}.
\end{bmatrix}
\end{align*}
Hence, the difference between sparse dictionary learning and problem (P) essentially lies in the underlying \textit{sparsity constraints}: while 
in the former one the only constraint is that sparse codes have no more than $p$ nonzero entries, in the latter they are further constrained \textit{to have at most one nonzero entry for each subdictionary}\footnote{Note that there are a few other differences. For example, in dictionary learning, atoms are usually unconstrained unit-norm vectors, while in our model they are constrained to be linear combinations of B-splines. Moreover, our problem involves an additional regularization function.}. Based on this analogy, we introduce in the next section a novel algorithm for efficiently approximating the solution of problem (P). 

\section{Learning algorithm}\label{sec:learning algorithm}

% The fitting problem (P) is challenging due to the $\ell_0$ norm constraint. Similarly to most dictionary learning algorithms, we resort to 

% Similarly to dictionary learning, the problem (P) is difficult to solve optimally. 
The problem of dictionary learning has proved challenging. In fact, even if the dictionary is  known, it can be NP-hard to represent a vector as a linear combination of the columns in the dictionary \cite{davis1997adaptive}. Problem (P) inherits the difficulty of dictionary learning, and we therefore propose an approximate algorithm that solves successively for the weights $\{ \boldsymbol\lambda_{j}^{(m)} \}$ and spline coefficients $\{ \mathbf{B}_j \}$. The proposed algorithm can be seen as an extension of the popular MOD algorithm \cite{engan1999method}, which alternates between sparse coding and dictionary updates.
% to our new sparsity model. 
%We present in the following sections the strategies employed to solve the two subproblems in our algorithm.

%We propose in this section an efficient approximate algorithm for solving (P). 
%In this section, we derive an algorithm to solve problem (P). This problem bears a suspicious resemblance with \textit{sparse dictionary learning}, where the goal is to simultaneously find the sparsest representations and the sparsifying dictionary atoms \cite{tosic2011dictionary}. Since the dictionary learning problem is NP-hard, 
%many algorithms consider an approximate solution that alternates between sparse coding and dictionary update. In a similar fashion, we propose an alternating approach that solves successively for the weights $\{ \boldsymbol\Lambda_j \}$ and spline coefficients $\{ \mathbf{B}_j \}$. We present in the following sections the strategies employed to solve the two subproblems.

\subsection{Weights update}
\label{sec:weights_update}
We assume that the spline coefficients matrices $\{ \mathbf{B}_j \}$ are given, and we define $\mbf{D}_j^{(m)} = \mathbf{S}_j^{(m)} \mbf{B}_j \in \mathbb{R}^{n \times L_j}$. We define the columns of each subdictionary $\mathbf{D}_j^{(m)} = \begin{bmatrix} \mbf{d}_{j, 1}^{(m)} & \dots & \mbf{d}_{j, L_j}^{(m)} \end{bmatrix}$ to be the \textit{atoms} of $\mathbf{D}_j^{(m)}$. Hence, an atom $\mbf{d}_{j,l}^{(m)}$ is obtained by applying the transfer function $f_{jl}$ to all the observations of the $j$th covariate: $\mathbf{d}_{j, l}^{(m)} = \begin{bmatrix} f_{jl} (x_{1j}^{(m)}) & \dots & f_{jl} (x_{nj}^{(m)}) \end{bmatrix}^T$. The weight estimation problem is given by
\begin{align*}
\min_{ \{ \boldsymbol\lambda_j^{(m)} \}_{j, m}}  \sum_{m=1}^N \left\| \mbf{y}^{(m)} - \sum_{j=1}^p \mathbf{D}_j^{(m)} \boldsymbol\lambda_{j}^{(m)} \right\|_2^2 \\
\text{ subject to }  \| \boldsymbol\lambda_{j}^{(m)} \|_0 \leq 1 \text{ and } \boldsymbol\lambda_j^{(m)} \geq \mbf{0} \text{ for all } j,m.
\end{align*}
The above problem can be seen as computing the best non-negative $p$-sparse approximations of the signals $\mbf{y}^{(m)}$ in the dictionary $\mbf{D}^{(m)} = [\mbf{D}_1^{(m)} | \dots | \mbf{D}_p^{(m)}]$, provided that no two active dictionary atoms belong to the same subdictionary. We first note that this problem is separable and therefore can be solved independently for each task. 
Next, we simplify the problem and drop the non-negativity constraints on $\boldsymbol\lambda_{j}^{(m)}$. Following the approach used in \cite{ekanadham2011sparse, fawzi2013classification}, non-negative coefficients can then be obtained in a post-processing step by including the negative of each atom in the dictionary\footnote{This post-processing step results in doubling the number of candidate transfer functions. That is, for covariate $j$, we have $2 L_j$ candidate transfer functions, as we consider the positive and negative versions of every (original) transfer function. Note also that this post-processing approach is an \textit{approximation} and is therefore not guaranteed to give the exact solution to the original (constrained) problem, with $2L_j$ transfer functions for each covariate $j$.}, as we have:
% Note that, in a post-processing step, one can obtain non-negative coefficients by augmenting the dictionary with negative versions of the atoms, as the following equality holds:
\begin{align*}
\sum_{j=1}^p \mathbf{D}_j^{(m)} \boldsymbol\lambda_j^{(m)} = \sum_{j=1}^p \begin{bmatrix} \mathbf{D}_j^{(m)} & - \mathbf{D}_j^{(m)} \end{bmatrix} \begin{bmatrix}
\max(0, \boldsymbol\lambda_{j}^{(m)}) \\
\max(0, -\boldsymbol\lambda_{j}^{(m)})
\end{bmatrix}.
\end{align*}
% As it is often done in sparse coding \cite{}, 
% Next, note that for any $\{ \boldsymbol\lambda_{j} ^{(m)} \}_{j=1}^p$, possibly with negative entries, we have:
%\begin{align*}
%\left\| \mbf{y}^{(i)} - \sum_{m=1}^p \mbf{D}_m \boldsymbol \lambda_{,m}^{(i)} \right\|_2^2 = \left\| \mbf{y}^{(i)} - \sum_{m=1}^{2p} \tilde{\mbf{D}}_m  \tilde{\boldsymbol \lambda}_{,m}^{(i)} \right\|_2^2,
%\end{align*}
%\begin{align*}
%& \left\| \mbf{y}^{(m)} - \sum_{j=1}^p \mbf{D}_j \boldsymbol \lambda_{j}^{(m)} \right\|_2^2 \\ = & \left\| \mbf{y}^{(m)} - \sum_{j=1}^{p} \begin{bmatrix}
%\mbf{D}_j  & -\mbf{D}_j
%\end{bmatrix}
%\begin{bmatrix}
%\max(0, \boldsymbol\lambda_{j}^{(m)}) \\
%\max(0, -\boldsymbol\lambda_{j}^{(m)})
%\end{bmatrix}
%\right\|_2^2 \\
%\triangleq  & \left\| \mbf{y}^{(m)} - \sum_{j=1}^p \tilde{\mbf{D}}_j \tilde{\boldsymbol\lambda}_{j}^{(m)} \right\|_2^2.
%\end{align*}
% Hence, by appropriately augmenting the dictionary with negative versions of the atoms, the non-negativity assumption on the coefficients $\tilde{\boldsymbol\lambda}_{j}^{(m)}$ is always satisfied. 
% We consequently drop the non-negativity constraint in the following.
For a single task, our weight estimation problem is written:
\begin{align*}
\min_{ \{ \boldsymbol\lambda_j \}_{j=1}^p}  \left\| \mbf{y} - \sum_{j=1}^p \mathbf{D}_j \boldsymbol\lambda_{j} \right\|_2^2 \\
\text{ subject to }  \| \boldsymbol\lambda_{j} \|_0 \leq 1 \text{ for all } j \in [p].
\end{align*}
To solve this problem, we propose the iterative algorithm Block Constrained Orthogonal Matching Pursuit (BC-OMP). It is an extension of the popular Orthogonal Matching Pursuit algorithm \cite{pati1993orthogonal, Mallat93} which is an efficient greedy method for solving sparse coding problems. At each iteration of the algorithm, we select the dictionary atom which has the strongest correlation with the residual, provided it belongs to an available subdictionary whose index is listed in $\mathcal{A}_{j-1}$.
The residual is then updated using an orthogonal projection onto the selected atoms. The availability set $\mathcal{A}_j$ is in turn updated to prevent selecting two atoms from the same subdictionary. Details of our approach are presented in Algorithm \ref{alg:algo_comp}.

\begin{algorithm}[t!]
 \caption{BC-OMP}
 \algorithmicrequire{\hspace{1mm}  Subdictionaries $\mbf{D}_1, \dots, \mbf{D}_p$, signal $\mbf{y}$.} \\
 \algorithmicensure{\hspace{1mm}  Weight vectors $\boldsymbol\lambda_1, \dots, \boldsymbol\lambda_p$.} \\
\label{alg:algo_comp}
\begin{algorithmic}
\STATE \textbf{Initialization:} \\ Available covariates: $\mathcal{A}_0 \leftarrow \{1, \dots, p\}$, residual: $\mbf{r}_0 \leftarrow \mbf{y}$, selected atoms: $\mbf{U}_0  \leftarrow \emptyset$, weight vectors: $\boldsymbol\lambda_j \leftarrow \mbf{0} \text{ for all } j \in [p]$.
%\[
%\begin{array}{l}
%\text{Available covariates:} \mathcal{A}_0 \leftarrow \{1, \dots, p\}, \\
%\text{Residual:} \mbf{r}_0 \leftarrow \mbf{y}, \\
%\text{Selected atoms:} \mbf{U}_0  \leftarrow \emptyset, \\
%\text{Weight vectors:} \boldsymbol\lambda_j \leftarrow \mbf{0} \text{ for all } j \in [p].
% \multicolumn{1}{c}{\boldsymbol\lambda_j \leftarrow \mbf{0} \text{ for all } j \in [p].} \\
%\end{array} 
%\]
\FORALL{$j=1,\dots,p$}
\STATE \textbf{Selection step:}
\begin{align*}
\{ k_j , l_j \} \leftarrow \argmax_{k \in \mathcal{A}_{j-1}} \argmax_{l \in \{1, \dots, L_k \}} \frac{\left| \scalprod{ \mbf{r}_{j-1} }{ \mbf{d}_{k, l} } \right|}{\| \mbf{d}_{k,l} \|_2}.
\end{align*}
\STATE \textbf{Update step:}
\[
\begin{array}{l}
\mathcal{A}_{j} \leftarrow \mathcal{A}_{j-1} \backslash \{ k_j\}, \mbf{U}_{j} \leftarrow \begin{bmatrix} \mbf{U}_{j-1} & \{ \mbf{d}_{k_j, l_j} \} \end{bmatrix} \\
\mbf{c}_{j}  \leftarrow \mbf{U}_{j}^\dagger \mbf{y}, \mbf{r}_{j}  \leftarrow \mbf{y} - \mbf{U}_{j} \mbf{c}_{j}.
\end{array}
\]
\ENDFOR
\FORALL{$j=1,\dots,p$}
\STATE Set $\boldsymbol\lambda_{k_j}[l_j] \leftarrow \mbf{c}_{p} [j]$.
\ENDFOR
\end{algorithmic}
\end{algorithm}

% Orthogonal Matching Pursuit (OMP) is an efficient greedy algorithm that solves the $p$-sparse approximation problem. 
%
%We therefore propose a modified version of Orthogonal 

% The main difference between the above problem and that of computing the best $p$-sparse approximation of signals $\mbf{Y}$ in a dictionary $\mbf{D} = [\mbf{D}_1 | \dots | \mbf{D}_p]$ formed by concatenating all the subdictionaries $\mbf{D_1}, \dots, \mbf{D_m}$ is the search space. 

% Letting $\mathbf{D}$ denote the concatenation of dictionaries $\{ \mbf{D}_m \}_{m=1}^p$, 
\subsection{Spline coefficients update}
We now solve the problem of learning the spline coefficients $\mbf{B}_j$ given the fixed weights $\boldsymbol\lambda_{j}^{(m)}$. % $\boldsymbol\Lambda_j$. 
We note that:
\begin{eqnarray*}
\sum_{j=1}^p \mathbf{S}_j^{(m)} \mathbf{B}_j \boldsymbol\lambda_j^{(m)}
& = & \sum_{j=1}^p ((\boldsymbol\lambda_j^{(m)})^T \otimes \mathbf{S}_j^{(m)}) \text{vec} (\mathbf{B}_j) \\
&& \hspace{-40mm} = \  \begin{bmatrix} 
(\boldsymbol\lambda_1^{(m)})^T \otimes \mathbf{S}_1^{(m)} & \dots & (\boldsymbol\lambda_p^{(m)})^T \otimes \mathbf{S}_p^{(m)}
\end{bmatrix}
\begin{bmatrix} 
\text{vec} (\mbf{B}_1) \\ \vdots \\ \text{vec} (\mbf{B}_p)
\end{bmatrix} \\ && \hspace{-40mm} \triangleq \mbf{Z}^{(m)} \mbf{b}.
\end{eqnarray*}
%\begin{eqnarray*}
%\text{vec} \left( \sum_{j=1}^p \mathbf{S}_j \mathbf{B}_j \boldsymbol\Lambda_j  \right)  
%& = & \sum_{j=1}^p (\boldsymbol\Lambda_j^T \otimes \mathbf{S}_j) \text{vec} (\mathbf{B}_j) \\
%&& \hspace{-35mm} = \  \begin{bmatrix} 
%\boldsymbol\Lambda_1^T \otimes \mathbf{S}_1 & \dots & \boldsymbol\Lambda_p^T \otimes \mathbf{S}_p
%\end{bmatrix}
%\begin{bmatrix} 
%\text{vec} (\mbf{B}_1) \\ \vdots \\ \text{vec} (\mbf{B}_p)
%\end{bmatrix} \triangleq \mbf{M} \mbf{b}.
%\end{eqnarray*}
Thus, the objective function becomes:
\begin{align*}
& \sum_{m=1}^N \| \mbf{y}^{(m)} - \mbf{Z}^{(m)} \mbf{b} \|_2^2 + \mbf{b}^T \boldsymbol\Sigma \mbf{b} = \| \text{vec} (\mbf{Y}) - \mbf{Z} \mbf{b} \|_2^2 + \mbf{b}^T \boldsymbol\Sigma \mbf{b},
\end{align*}
with $\text{vec} (\mbf{Y}) = \begin{bmatrix} \mbf{y}^{(1)} \\ \vdots \\ \mbf{y}^{(N)} \end{bmatrix}$ and $\mbf{Z} = \begin{bmatrix} \mbf{Z}^{(1)} \\ \vdots \\ \mbf{Z}^{(N)} \end{bmatrix}$. The minimum of the above least-squares program with respect to $\mbf{b}$ is given by:
\begin{align}
\label{eq:closed_form_spline_update}
\hat{\mbf{b}} = (\mbf{Z^T Z} + \boldsymbol\Sigma)^{-1} \mbf{Z^T} \text{vec} (\mbf{Y}).
\end{align}
Given the spline basis coefficients $\mathbf{B}_j$, the transfer functions can then be obtained by multiplying the obtained coefficients with the spline basis vectors (Eq. (\ref{eq:spline_tfs})).
% Thus we obtain:
% \begin{align*}
% \min_{\mbf{b}} \| \text{vec} (\mbf{Y}) - \mbf{M} \mbf{b} \|_2^2 + \mbf{b^T} \boldsymbol\Sigma \mbf{b},
% \end{align*}
% and the solution to this least-squares problem is:

% A fitting algorithm  (\ref{eq:model_shared_matrixForm}) and derive an algorithm that leverages the recent work in sparse dictionary learning.

% in this section the learning of transfer functions, etc...

\subsection{Complete learning algorithm}

The complete algorithm is shown in Algorithm \ref{alg:algo_clustering}. Using a random initialization of the spline coefficients, we iterate through the weights update and spline coefficients update steps, until a termination criterion is met. In this paper, we terminate the algorithm after a fixed number of iterations. In a final step, the matrices $\mathbf{B}_j$ and $\boldsymbol\lambda_j^{(m)}$ are modified to ensure the non-negativity of the weights, as discussed in Section \ref{sec:weights_update}.

\begin{algorithm}[ht]
 \caption{Multi-task additive model algorithm}
 \algorithmicrequire{\hspace{1mm}  Covariates $\{x_{ij}^{(m)}\}_{i,j,m}$, response variables $\{y_i^{(m)}\}_{i,m}$, parameters $L_1, \dots, L_p$ and $\nu$.} \\
 \algorithmicensure{\hspace{1mm}  Spline coefficients $\{ \mathbf{B}_j \}_j$, scaling weights $\{ \boldsymbol\lambda_j^{(m)} \}_{j,m}$. } \\
\label{alg:algo_clustering}
\begin{algorithmic}
\STATE Initialize $\mbf{B}_1, \dots, \mbf{B}_p$ with random entries from $\mathcal{N} (0,1)$.
\WHILE{not converged}
\STATE \textbf{Weights update:} Use BC-OMP for each response variable $\mbf{y}^{(m)}$ to estimate  $\{ \boldsymbol\lambda_j^{(m)} \}_{j\in[p], m\in[N]}$. 
\STATE \textbf{Spline coefficients update:}  Use Eq. (\ref{eq:closed_form_spline_update}) to update the spline coefficients.
%\STATE \textbf{Normalize the columns of } $\{ \mathbf{B}_j \}$: For all $j,l,m$:
%\begin{align*}
%\boldsymbol\beta_{jl} & \leftarrow \frac{\boldsymbol\beta_{jl}}{\| \mbf{S}_j \boldsymbol\beta_{jl} \|_2}, \\
% \lambda_{jl}^{(m)} & \leftarrow \| \mbf{S}_j \boldsymbol\beta_{jl} \|_2 \lambda_{jl}^{(m)}.
%\end{align*}
\ENDWHILE
\STATE \textbf{Ensure the non-negativity of the weights:}
\begin{align*}
\mathbf{B}_j & \leftarrow \begin{bmatrix} \mathbf{B}_j & -\mathbf{B}_j \end{bmatrix}, \\
\boldsymbol\lambda_j^{(m)} & \leftarrow \begin{bmatrix} \max(0, \boldsymbol\lambda_j^{(m)}) \\ \max(0, -\boldsymbol\lambda_j^{(m)})
\end{bmatrix},
\end{align*}
for all $j, m$.
\end{algorithmic}
\end{algorithm}

% \subsection{Computational complexity}

We briefly analyze the computational complexity of Algorithm \ref{alg:algo_clustering}. We assume for simplicity that $L_j = L$ for all $j \in [p]$. The BC-OMP algorithm involves a selection and an update step whose complexity are $O(Lpn)$ and $O(pn + p^2)$ respectively. Assuming that $p<n$, the selection step dominates and the overall complexity of BC-OMP is $O(L p^2 n)$. Doing this operation for each task results in a complexity of $O(n N L p^2)$.
 The spline coefficients update involves the computation in Eq. (\ref{eq:closed_form_spline_update}). To compute the complexity of this operation, note that $\mbf{Z}$ has $nN$ rows and $LTp$ columns (where $T$ is the number of spline basis functions, or equivalently the number of columns of $\mbf{S}_j^{(m)}$, assumed to be equal for all covariates $j$ for simplicity). For typical problems, this matrix is tall and the complexity is driven by the computation of $\mbf{Z}^T \mbf{Z}$, which is of complexity $O(n N (LTp)^2)$. Hence, assuming that we run Algorithm \ref{alg:algo_clustering} for a fixed number of iterations, the complexity of our overall algorithm is $O(n N (LTp)^2)$.
 
Our algorithm is therefore \textit{linear} in the number of tasks $N$, and dimension $n$, while being quadratic with respect to the number of candidate transfer functions per task $L$. In comparison, learning an additive model independently for each task has complexity $O(nN (Tp)^2)$. Compared to the independent additive model approach, the price to pay of our algorithm is therefore $L^2$, which remains small in most problems of interest.
% Note that the complexity of learning independently one additive model per task is \textit{cubic} with respect to the number of tasks; our algorithm therefore provides the opportunity to scale additive models to much larger tasks, where traditional additive models cannot be directly applied.
% Learning independently one additive model per task is therefore pr

% \red{Take care of the fact that the subdictionaries are of coherence 1, because you replicate with the sign... Handle that properly..}

\section{Recovery condition for BC-OMP}
\label{sec:recovery_cond_COMP}
% \red{Mention in a place here the other papers that have recovery guarantees on OMP block-wise: Eldar's paper, and Lorenzo Peotta. Say that we are fundamentally different here.}

We analyze in this section the weights update algorithm BC-OMP. While BC-OMP represents one building block of the global algorithm (Algorithm \ref{alg:algo_clustering}), an analysis of the recovery conditions of BC-OMP is important as it provides insights onto the success of our algorithm. In addition, our recovery condition also applies to OMP and is interesting in its own right.

% While the success of C-OMP does not guarantee the success of the complete learning algorithm (Algorithm \ref{alg:algo_clustering}), C-OMP constitutes a crucial part of our algorithm, and an analysis of the recovery conditions is important as it provides insights onto the success of our algorithm. In addition, it provides new recovery conditions for OMP that are interesting in its own right.
% While a successful recovery of the weights does not guarantee that the success of the whole learning algorithm in Algorithm \ref{}, a correct recovery of the weights is important to ensure 

We suppose that $\mbf{y}$ is a superposition of $p$ elements in $\mbf{D} = [\mbf{D}_1 | \dots | \mbf{D}_p]$ such that no two active elements belong to the same subdictionary $\mbf{D}_j$, i.e., $\mbf{y} = \sum_{j=1}^p \gamma_j \mbf{d}_{j, l_j}$. For simplicity, we further assume that the atoms $\mbf{d}_{j, l_j}$ are linearly independent and the $\gamma_j$ are all nonzero\footnote{Otherwise, the signal has a representation with fewer atoms and one can remove the unused covariates $j \in [p]$.}. We develop a sufficient condition for the recovery of the correct atoms using BC-OMP.
% First, observe that our signal model is included in the set of exactly $p$-sparse signals, which allows any combination of $p$ atoms in $\mbf{D}$.
% The main difference between this model and the set of exactly $p$-sparse signals is that we disallow two atoms of the same subdictionary to take part in the decomposition. Our signal model therefore forms a subset of the set of $p$-sparse signals.

% While our signal model is included in the set of $p$-sparse signals, the main difference 

% The main difference between our signal model and the traditional $p$-sparse signal model \cite{tropp2004greed} lies in the 

We first note that the difference between OMP and BC-OMP algorithms lies in their search space: while OMP selects atoms from the dictionary $\mathbf{D}$ having maximal inner product with the residual, BC-OMP further imposes a constraint that the selected atom belong to an available subdictionary where no atoms have been previously selected. It follows that if OMP succeeds in the recovery of the correct atoms of $\mathbf{y}$, the same holds for BC-OMP.
% We first note that if OMP succeeds in the recovery of the correct atoms, then the same holds for C-OMP. 
Therefore, any condition that guarantees the recovery of OMP is \textit{a fortiori} a recovery condition for BC-OMP. Many OMP recovery conditions have been proposed in the literature (see e.g., \cite{tropp2004greed, davenport2010analysis}). The following theorem in \cite{tropp2004greed} gives a popular and practical recovery condition of OMP for the global dictionary $\mbf{D}$:
\begin{theorem}[\cite{tropp2004greed}]
\label{theorem:recovery_condition_OMP} Let
\begin{eqnarray*}
\mu &\triangleq& \displaystyle\max_{\substack{\mbf{d}, \mbf{d}' \in \mbf{D} \\ \mathbf{d} \neq \mathbf{d}'}} |\scalprod{\mbf{d}}{\mbf{d}'}|. 
\end{eqnarray*}
OMP recovers every superposition of $p$ atoms from $\mbf{D}$ whenever the following condition is satisfied:
\begin{align}
\label{eq:recovery_condition_OMP}
p < \frac{1}{2} \left( \mu^{-1} + 1 \right).
\end{align}
\end{theorem}

The quantity $\mu$, called \textit{coherence}, measures the similarity between dictionary atoms. 
The values of $\mu$ that are close to $1$ may violate the recovery condition in Eq.~(\ref{eq:recovery_condition_OMP}), thus 
leaving us without any guarantee that OMP or BC-OMP will recover the correct atoms.
Unfortunately enough, in our multi-task learning framework, $\mu$ is typically close to $1$. To see this, note that the inner product between two atoms of the same subdictionary is equal to $\scalprod{\mbf{d}_{j,l}}{\mbf{d}_{j,l'}} = \sum_{i=1}^n f_{jl} (x_{ij}) f_{jl'} (x_{ij})$. In practice, transfer functions in the same subdictionary often bear strong resemblance, e.g., similar monotic behavior (see Sec.~\ref{sec:experiments} for examples). Thus, $f_{jl} \approx f_{jl'}$ and $|\scalprod{\mbf{d}_{j,l}}{\mbf{d}_{j,l'}}| \approx 1$, which leads to a large coherence value. On the other hand, the inner product of atoms from different subdictionaries $|\scalprod{\mbf{d}_{j,l}}{\mbf{d}_{j',l'}}| = \sum_{i=1}^n f_{jl} (x_{ij}) f_{j'l'} (x_{ij'})$ is close to zero when the covariates $j$ and $j'$ are ``sufficiently independent''.\footnote{More precisely, if we model the covariates as random variables $X_j$ and $X_{j'}$, then $1/n \sum_{i=1}^n f_{jl} (x_{ij}) f_{j'l'} (x_{ij'})$ can be seen as a sample estimate of the population covariance $\mathbb{E} [ f_{jl} (X_j) f_{j'l'} (X_{j'}) ]$ which will be $0$ if $X_{j}$ and $X_{j'}$ are independent. Note that in this argument we use the fact that the transfer functions are centered (see Sec. \ref{sec:additive_model_review}).} To circumvent the above violation of the recovery condition in Theorem \ref{theorem:recovery_condition_OMP} due to the large \textit{global coherence} of the dictionary, we first define coherence within and across subdictionaries: % new \textit{within} and \textit{across} the subdictionaries that distinguish between atoms from the same subdictionary, and atoms from different subdictionaries.

\begin{definition}\label{def:our_coherence}
\begin{align*}
\mu_{intra}& \triangleq \max_{j \in [p]} \max_{\substack{(l, l') \in [L_j] \\ l \neq l'}}\left| \scalprod{\mbf{d}_{j,l}}{\mbf{d}_{j, l'}} \right|, \\
\mu_{inter} & \triangleq \max_{\substack{(j, j') \in [p] \\ j \neq j'}} \max_{(l, l') \in [L_j] \times [L_{j'}]} \left| \scalprod{\mbf{d}_{j, l}}{\mbf{d}_{j', l'}} \right|.
\end{align*}
\end{definition}

%  While $\mu_{intra}$ is expected to be close to one, $\mu_{inter}$ essentially depends on the correlation between the covariates, and therefore typically takes small values. 
% We have the following recovery condition for C-OMP:

Using these definitions, we derive the following recovery condition for BC-OMP:
\begin{theorem}[Recovery condition]
\label{theorem:our_recovery_condition}
If the following condition holds:
\begin{align*}
\mu_{intra} + 2 (p-1) \mu_{inter} < 1,
\end{align*}
then BC-OMP recovers the correct atoms and their coefficients.
\end{theorem}

The detailed proof can be found in the Appendix. Unlike the recovery condition in Theorem \ref{theorem:recovery_condition_OMP}, Theorem \ref{theorem:our_recovery_condition} does not depend on the global coherence of the dictionary. This recovery condition is particularly interesting in our applications where $\mu_{intra}$ typically takes large values due to strong resemblance between the transfer functions in the same subdictionary, while $\mu_{inter}$ is small when the covariates are sufficiently statistically independent. % as the covariates are statistically almost independent.
% TO PUT SECTION D'APRES
% For the purpose of illustration, we have $\mu_{intra} = 0.92$ and $\mu_{inter} = 0.01$ in the example of Fig. \ref{fig:correlation_tfs}. Our recovery condition is therefore satisfied, while the condition of Theorem \ref{theorem:recovery_condition_OMP} is clearly violated.
% END TO PUT SECTION D'APRES
Interestingly, Theorem \ref{theorem:our_recovery_condition} shows that, in the special case where subdictionaries are orthogonal to each other, the parameter $\mu_{intra}$ can take values arbitrarily close (but not equal) to $1$ and BC-OMP still succeeds in the recovery. In contrast, the recovery condition of Theorem \ref{theorem:recovery_condition_OMP} is not satisfied in this case since the global coherence $\mu$ is close to $1$. Note that our recovery condition is not limited to BC-OMP but also valid for the OMP algorithm, hence it extends existing sparse representation theory for an interesting class of sparsity structure.
%SHORT VERSION
%We conclude this section by putting our theoretical analysis into more context. In \cite{peotta2007matching}, a sufficient condition is given for the recovery of atoms from correct subdictionaries when the dictionary is built from an incoherent union of coherent subdictionaries. This is different from our condition that guarantees recovery of the exact support when the signal contains at most one atom per subdictionary. The authors of \cite{eldar2010block} introduce a model that constrains sparse codes to be active in blocks and show recovery conditions for a block version of OMP. Again, the sparse model we consider in this paper is different as it does not constrain non-zero entries to appear in blocks. 
% END SHORT VERSION

Finally, we draw the reader's attention to some results related to the proposed recovery condition in Theorem \ref{theorem:our_recovery_condition}. In \cite{peotta2007matching}, the authors provide a new analysis of Matching Pursuit, when the dictionary is built from an incoherent union of possibly coherent subdictionaries. A sufficient condition that guarantees the selection of atoms from the correct subdictionaries is shown. In other words, the exact recovery of atoms is dropped and a sufficient condition for the weaker subdictionary recovery property is shown. This is completely different from our setting, where we require the correct atoms to be recovered when the signal contains at most one atom per subdictionary. In \cite{eldar2010block}, the ``block sparse'' model is introduced: the signals' non-zero entries appear in blocks rather than being spread throughout the vector. Coherence-based recovery conditions for a block version of OMP are shown. Once again, our model significantly differs from this one, as we assume \textit{one} active component per subdictionary (or block), whereas the work of \cite{eldar2010block} assumes that nonzero entries occur in clusters.

\section{Experimental results}\label{sec:experiments}

In this section, we present experimental results. Baseline methods and practical implementation details of our algorithm are explained in Sec.~\ref{sec:experimental_setup}. Sec.~\ref{sec:synthetic_experiment} reports results on synthetic data, and the following two sections show results on electric load forecasting problems. % Sec.~\ref{sec:prediction_smart_meters} we report results from experiments
% with smart meter data. 
% In the supplementary material accompanying this submission, we also apply our method to the examination of intra-daily temperature effects in aggregated electricity demand data. % , and in Sec.~\ref{sec:intra_signal_additive} we apply our method to the examination of intra-daily temperature effects in aggregated electricity demand data. 
More background on using additive models for electricity demand forecasting can be found in \cite{ba2012adaptive, fan2012short} for example.

%  implementation details related to the proposed method in Section ... . Then, we present etc...

\begin{figure*}[t]
\centering
\subfigure[]{
\includegraphics[width=0.3\textwidth]{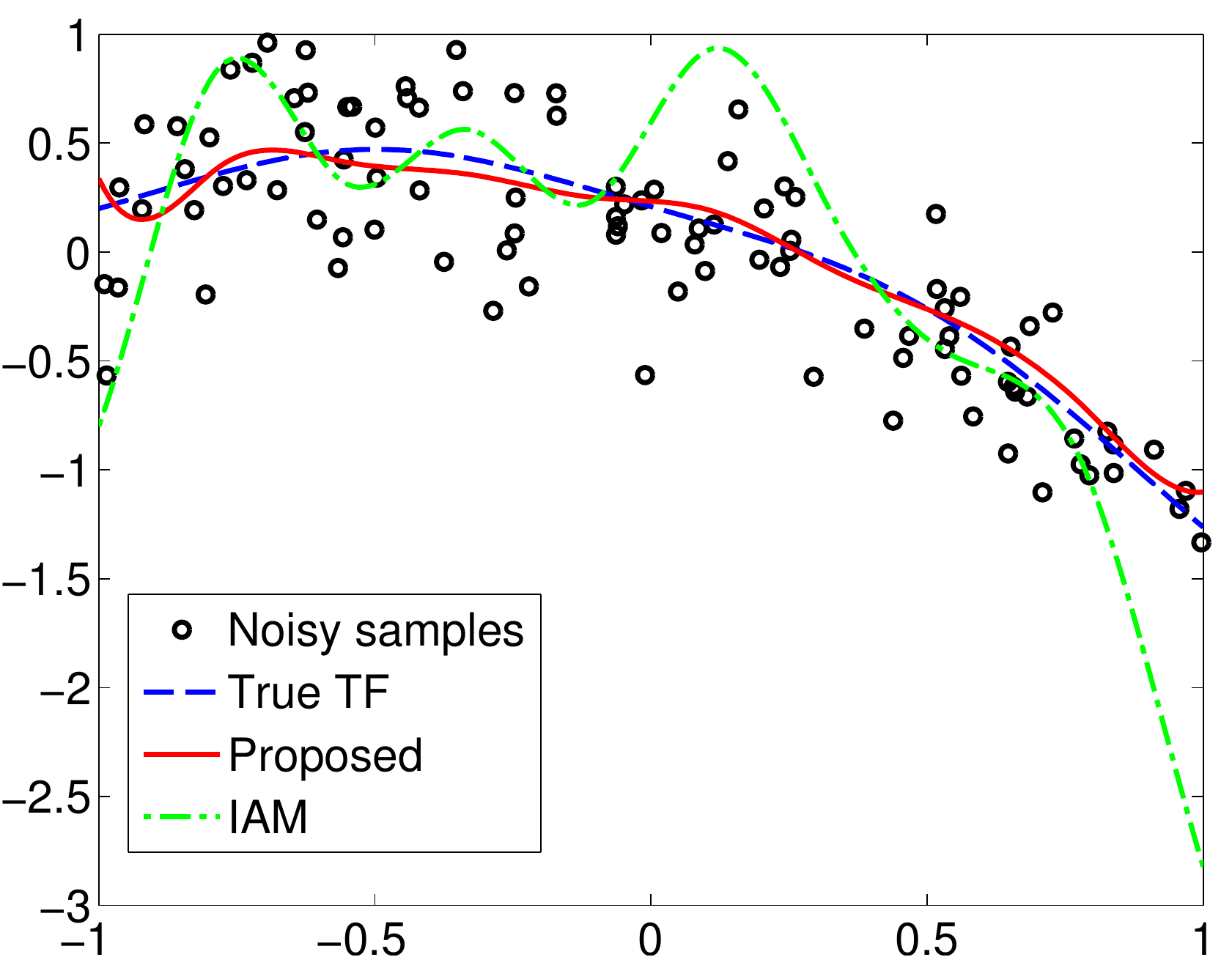}
}
\centering
\subfigure[]{
\includegraphics[width=0.3\textwidth]{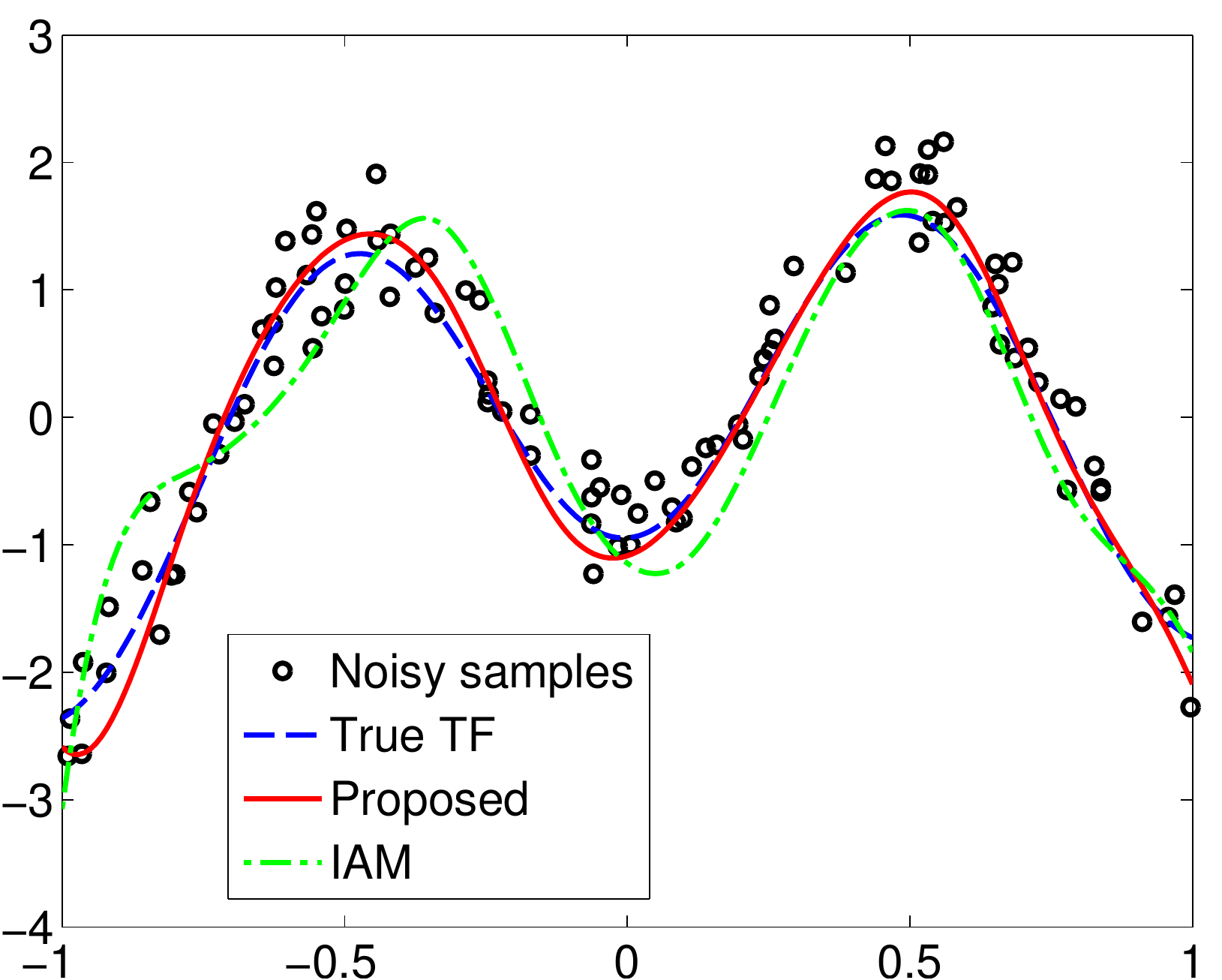}
}
\centering
\subfigure[]{
\includegraphics[width=0.3\textwidth]{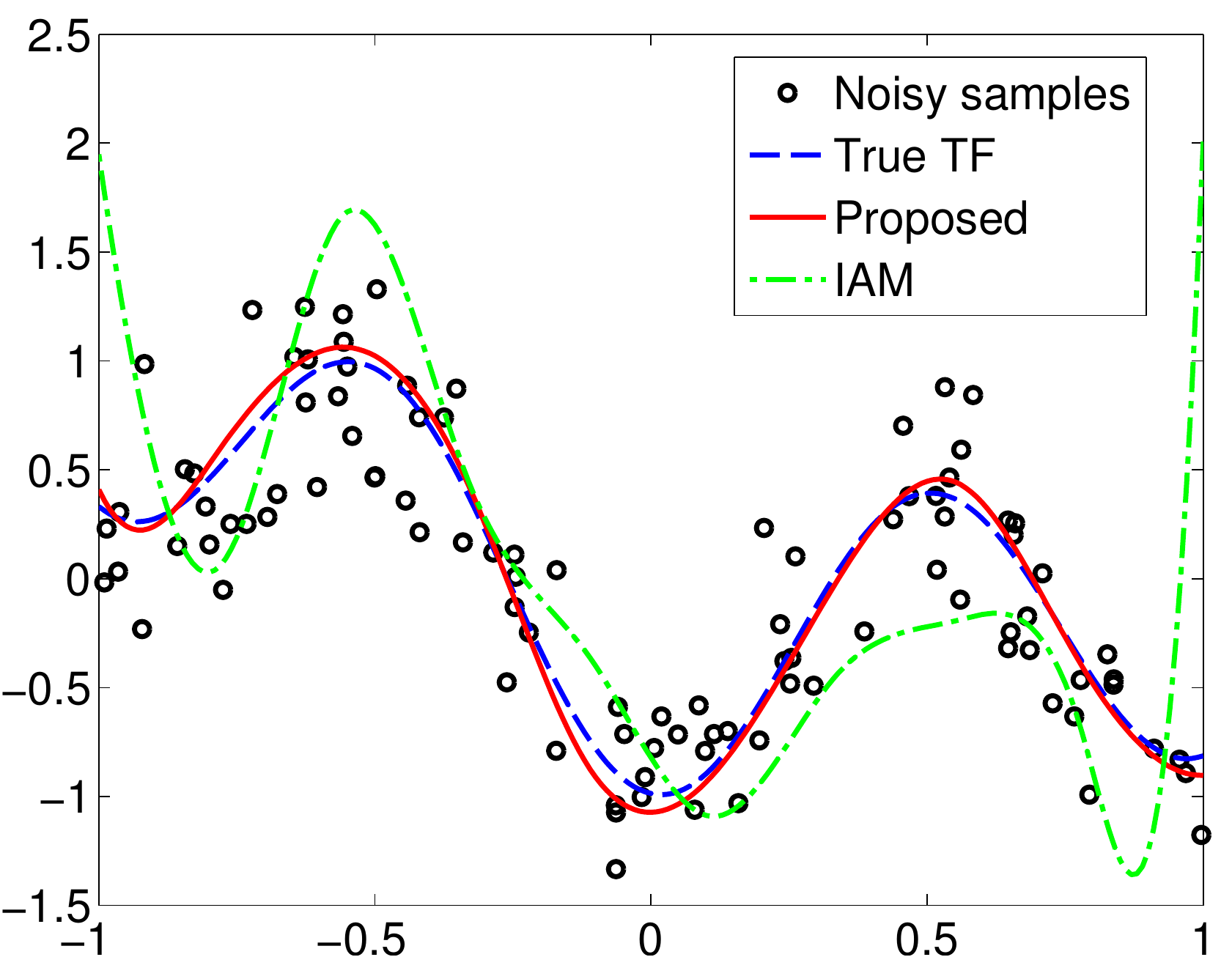}
}
\caption{\label{fig:toy_example} Estimated transfer functions for the synthetic experiment, using the proposed and IAM methods. The black dots indicate the noisy observations used to estimate the transfer functions, and the blue dashed lines represent the \textit{true} transfer functions. The results are shown for a \textit{fixed covariate}, leading to the $L=3$ estimated transfer functions that are illustrated. % True and estimated transfer functions using our approach, and IAM, for a fixed covariate. 
For the IAM method, the depicted transfer functions correspond to estimations obtained on arbitrary tasks involving the true transfer functions.}
\end{figure*}

\subsection{Experimental setup}
\label{sec:experimental_setup}
We compare the proposed multi-task learning approach to the following baseline regression methods:

% \noindent \textbf{Linear Regression (LR):} A linear regressor is learned independently for each task.
% using $\epsilon$-SVR. The penalty parameter $C$ is set using a cross-validation procedure for each task.
\begin{enumerate}
\item \noindent \textbf{Linear Regression (LR):} A linear regressor is learned independently using $\epsilon$-SVR \cite{drucker1997support} with a linear kernel for each task. The penalty parameter $C$ is set using a cross-validation procedure for each task. We use the Liblinear implementation \cite{fan2008liblinear}.

\item \noindent \textbf{Support Vector Regression with Radial Basis Function kernel (SVR-RBF): } We learn an $\epsilon$-SVR-RBF 
regressor independently for each task where the penalty parameter $C$ and kernel bandwidth $\sigma$ are determined using cross-validation. We use the LibSVM implementation \cite{CC01a}.

\item \noindent \textbf{Independent Additive Models (IAM):} An additive model is fitted for each task independently using
the \texttt{mgcv} package in \texttt{R} \cite{wood2006generalized}.
%\footnote{http://cran.r-project.org/web/packages/mgcv/mgcv.pdf}

\item \noindent \textbf{K-Means and Additive Model (KAM):} This is a two-step approach, where in the first step we use the K-means algorithm to group the set of tasks into different clusters. In the second step, one additive model is learned independently for each cluster centroid. The prediction of a signal is given by the prediction for the centroid of the cluster it belongs to. 

\end{enumerate}
% \red{To discuss the choice of the parameters here...}

% In our method, we use a B-spline representai

%  \red{Mention that while we use spline basis for modeling Time Of Day and Time Of Year covariates, we use a Dirac basis for DayType}

We now discuss practical aspects of our algorithm. For simplicity, we have chosen the regularization value $\nu = 1$ in all our experiments (see Eq.~\ref{eq:regularization}). Note that a cross-validation procedure is likely to give better results, but is more computationally expensive. 
Moreover, in all experiments we set the parameters $L_j = L$ for all covariates $j \in [p]$. 
We envision that, in real-world applications, the parameters $L_j$ will be manually selected by domain experts (possibly in an iterative procedure) to find the optimal trade-off between predictive performance and model interpretability from their
point of view.
\iffalse
When the goal is to maximize the predictive performance, these parameters can be set through a cross-validation procedure (\red{or using the same techniques employed by K-means (BIC, AIC,...) ask someone who knows more about that stuff}). 
\fi
In the experiments below, we show results for different values of $L$ to evaluate how the choice of $L_j$ affects the predictive performance.
Finally, similarly to K-means, our proposed algorithm can incur the problem of ``empty clusters'' when the number $L$ of transfer functions per covariate is large. We circumvent this problem by checking at each iteration for unused transfer functions and, when such a transfer function is detected, replacing it with the transfer function that leads to minimum error for the task with the currently highest approximation error.

%
%\begin{itemize}
%\item How do you fix the parameter $\nu$?
%\item The parameters $L_j$, that control the number of transfer functions, is also a crucial parameter to set. In a practical application, this parameter is typically tuned by a domain expert as keeping these parameters low is important for the interpretability of the model. In the experiments, we set all $L_i = L$ to be equal.
%\item If in one of the iterations, a cluster is empty, we replace it with one that minimizes the error (to discuss that in detail).
%\end{itemize}

% We now discuss the choice of the parameters. The parameter $\nu$ controls the importance of the regularization term in our learning formulation. In the experiments, we set this parameter by cross-validation in the set $...$. The parameters $L_m$, that control the number of transfer functions, is also a crucial parameter to set. In a practical application, this parameter is typically tuned by a domain expert as keeping these parameters low is important for the interpretability of the model. In the experiments, we set all $L_i$ to be equal, and ..

%\begin{itemize}
%\item How to set the hyper-parameters?
%\item If there is less than $L$ clusters, replace them with the one that minimizes the error.
%\end{itemize}

% \subsection{Application: Electricity load forecasting}

% \subsubsection{Inter-signal additive model clustering}

\subsection{Synthetic experiment}
\label{sec:synthetic_experiment}

In our first experiment, we generate $n=100$ samples according to the multi-task additive model in Eq. (\ref{eq:model_shared}), with $p=10$ covariates, $N=200$ tasks and $L_1 = \dots = L_p = L = 3$ candidate transfer functions per covariate. For simplicity, we take the covariates $x_{ij}^{(m)}$ to be equal for all tasks (i.e., $x_{ij}^{(m)} = x_{ij}$ for all $m \in [N]$), and randomly sample $x_{ij}$ from the uniform distribution in $[-1, 1]$. In this synthetic example, the ground truth transfer functions $f_{jl}$ are randomly generated smooth functions, and the scaling weights are chosen to be non-negative random numbers. Finally, the model noise $\epsilon_i^{(m)}$ is iid, and follows the standard normal distribution.

\begin{figure}[t]
\centering
\includegraphics[width=0.35\textwidth]{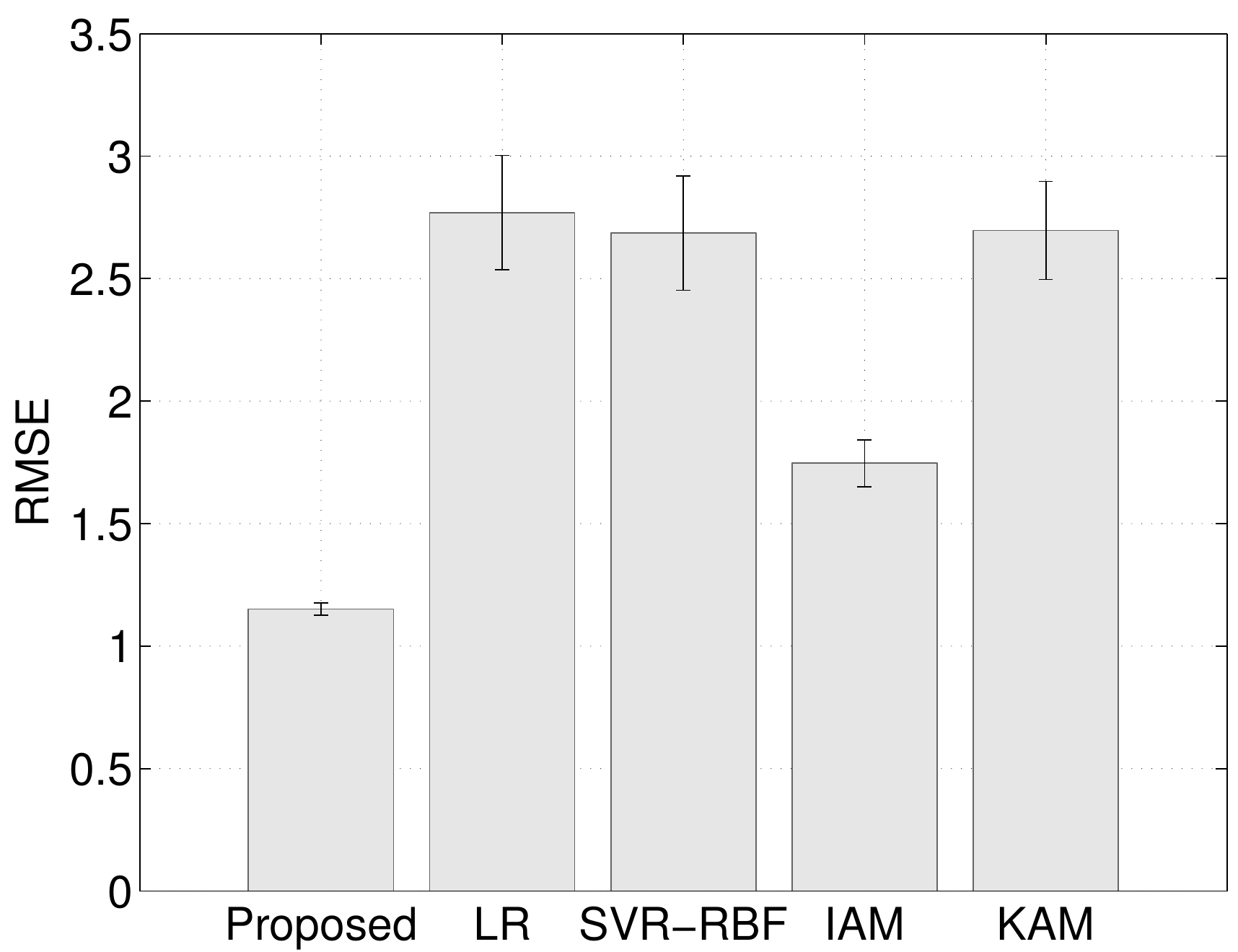}
\caption{\label{fig:rmse_synthetic} RMSE on test data of the proposed method, and the competing methods for the synthetic experiment. The experiments are performed on $50$ independent trials.}
\end{figure}

We first assess the quality of the estimated transfer functions using our algorithm, and compare it to the ground truth transfer functions (known in this synthetic setting), as well as the functions estimated with IAM method (which treats each task independently). For a fixed covariate, we show in Fig. (\ref{fig:toy_example}) (a-c) the $L = 3$ associated transfer functions, as well as the proposed and IAM estimations.
% The results are shown in Fig. (\ref{fig:toy_example}) (a-c), and depict the $L=3$ true transfer functions
% Fig. (\ref{fig:toy_example}) (a-c) illustrates the $L=3$ true transfer functions (all corresponding to one covariate). 
Clearly, the estimation of the true transfer functions using our multi-task approach is much more accurate and resilient to noise than IAM. In fact, the true and estimated functions using our approach nearly coincide, despite the fact that observations are highly noisy, and the relatively low sample size. We then compare the prediction performance (in terms of Root Mean Squared Error -- RMSE) of the proposed method to other competitor methods on a test set of $400$ samples. The results are illustrated in Fig. \ref{fig:rmse_synthetic}.
% In Fig. \ref{fig:toy_example} (b), our algorithm is compared to other methods in terms of Root Mean Squared Error (RMSE) on a test set of $400$ samples. 
The proposed approach leads to significantly lower RMSE compared to other approaches on this synthetic example. Despite the limited number of training data, our approach correctly detects and leverages the correlation between the different tasks to significantly improve the results.
% , and the scaling weights $\lambda_{jl}^{(m)}$ are chosen iid at random.
% randomly $p=10$ covariates in the interval $[-1,1]$

\subsection{Modeling of smart meter data}
\label{sec:prediction_smart_meters}

%\begin{figure}[t!]
%\centering
%\subfigure[]{
%\includegraphics[width=0.2\textwidth]{correlation_atoms.pdf}
%}
%\centering
%\subfigure[]{
%\includegraphics[width=0.2\textwidth]{transfer_functions_ToD.pdf}
%}
%\caption{\label{fig:correlation_tfs} Left: Correlation between the different atoms in a dictionary ($L_1 = L_2 = L_3 = 5$). Right: Two ``Time of Day'' transfer functions that are highly similar, resulting in a high coherence of subdictionaries.}
%\end{figure}
\renewcommand{\arraystretch}{1.1}
\begin{table*}
  \caption{\label{table:results_CER} Average RMSE over all $4,066$ tasks in the CER data set, and model complexity. For the proposed and KAM method, the results with $L = 5$ are shown. For easier comparison, the fourth column shows the \textit{normalized} model complexity, i.e., the model complexity divided by $Np$ where $N$ is the number of tasks and $p$ the number of covariates. $T$ is the number of elements in the spline basis (equal for all covariates, for simplicity).}  % \colored{For our method; 20 iterations, 3 restarts (taking the best on training), smoothness and transfer parameters equal to $1$.} 
  \label{tab:ISONEresults}
  \centering
  \begin{tabular}{@{}lrrrrrrrrrrrr@{}}
  \toprule
  Method & \multicolumn{2}{c}{RMSE} & \phantom{abc} & \multicolumn{2}{c}{Model complexity} \\ \cmidrule{2-3} \cmidrule{5-6}
   & Training & Testing && Theoretical & Numerical example \\ \midrule
Proposed & $2.6$ & $2.7$ && $p(TL + 2N)$ & $2.01$ \\ 
   \\
   LR & $3.1$ & $3.1$ && $Np$ & $1$ \\
   \\
   SVR-RBF & $\mathbf{2.2}$ & $\mathbf{2.5}$ && $Nn(p+1)$ & $\geq 3800$ \\
   \\
   IAM & $2.3$ & $2.6$ && $pTN$ & $12$ \\ \\
   KAM & $3.1$ & $3.2$ && $pTL$ & $\mathbf{0.01}$
   \\ \bottomrule
  \end{tabular}
\end{table*}
In this experiment we use data from a smart metering trial of the Irish Commission for Energy Regulation (CER) \cite{cer_dataset_alternative}.
The data set contains half-hourly electricity consumption data from July 14, $2009$ to December 31, $2010$
for approx.~$5,000$ residential (RES) and small-to-medium enterprise (SME) customers. 
It comes with survey information about different demographic and socio-economic indicator, e.g., number of people living in the household, type of appliances, and business opening times.
In our experiment, we only use customers that do not have any missing consumption data, leaving us with a total
of $N = 4,066$ meters out of which $3,639$ are residential and $427$ SME customers. 
Since half-hourly smart meter data is very volatile due to the stochastic nature of electricity consumption at the individual household level, we aggregate each signal over $6$ time points to obtain one measurement every $3$ hours.
We split the data into $12$ months of training and $6$ months of test data. We consider a simple additive model with ``Hour Of Day'', ``Time Of Year'' and ``Day of Week'' as covariates. % We use $L=K=5$ in our proposed method and the KAM algorithm.
Table \ref{table:results_CER} shows the average RMSE on the training and test data over all $N = 4,066$ meters. Here, we use $L = 5$ for the number of candidate transfer functions per covariate in our approach, and the number of clusters in the KAM method.
In terms of predictive performance, our proposed approach clearly outperforms LR and KAM;
it performs only slight worse than IAM albeit using only $L=5$ different transfer functions per covariate while IAM
learns independently one additive model per signal. Note that the methods based on additive models are
competitive with SVR-RBF, while the latter approach is computationally expensive at training and test time, which makes it only moderately suitable for large-scale problems, besides leading to models that are unfortunately difficult to interpret.

In an attempt to quantitatively compare the interpretability of the different methods, Table \ref{table:results_CER} shows the \textit{model complexity} of the different methods learned on the CER data, defined as the number of scalar variables needed to store the model\footnote{For the SVR-RBF method, it is assumed for simplicity that the number of support vectors is equal to the number of training points $n$.}. For easier numerical comparison we divide the numbers by $Np$, i.e., the number of tasks times the number of covariates.
As it can be seen, the SVR-RBF is the most complex model, since it depends on the number of observations $n$. IAM also has a high complexity as it fits one additive model per task. On the other hand, our proposed approach only needs $pTL$ variables for the transfer functions, and $2Np$ values to store the weights $\{\boldsymbol\lambda_j^{(m)}\}$. 
In the experiment on the CER data, this results in a complexity of $2.01$, where $0.01$ results from the representation of the transfer functions, and $2$ from storing the weight coefficients for each task. We conclude that our proposed approach finds a good trade-off between predictive performance and model complexity. 

% We therefore conclude that our approach therefore achieves therefore a good 
% In terms of normalized complexity on CER, the first part amounts for approximately $0.01$, while the coefficient part has a normalized complexity of $2$. 
% The drawback of the latter is however the lack of interpretability of the learned parameters. 

\begin{figure}[t]
\centering
\includegraphics[width=0.3\textwidth]{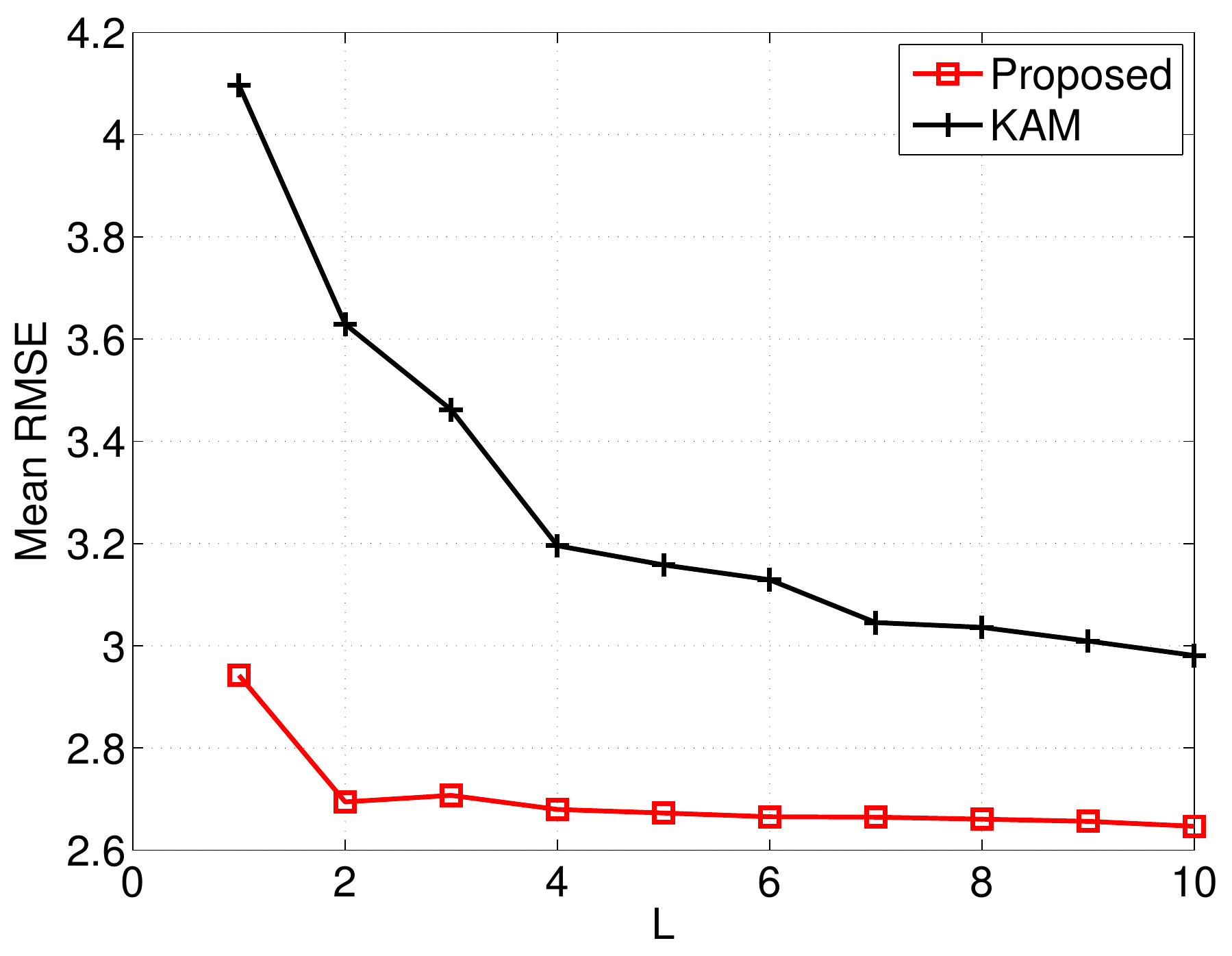}
\caption{\label{fig:rmse_vs_L} Average RMSE on the CER data set vs.~number of candidate functions (per covariate) and clusters $L$, respectively, for the proposed and the KAM method.}
\end{figure}

Fig.~\ref{fig:rmse_vs_L} shows the predictive performance of the proposed algorithm and the KAM method for different values of
$L$. One can see that, already for values of $L\geq2$, our method reaches an accuracy that is close to the performance of independently learned additive models. The performance of KAM is consistently worse, regardless of the number of clusters.

\begin{figure*}[ht]
\centering
\subfigure[]{
\includegraphics[width=0.2\textwidth]{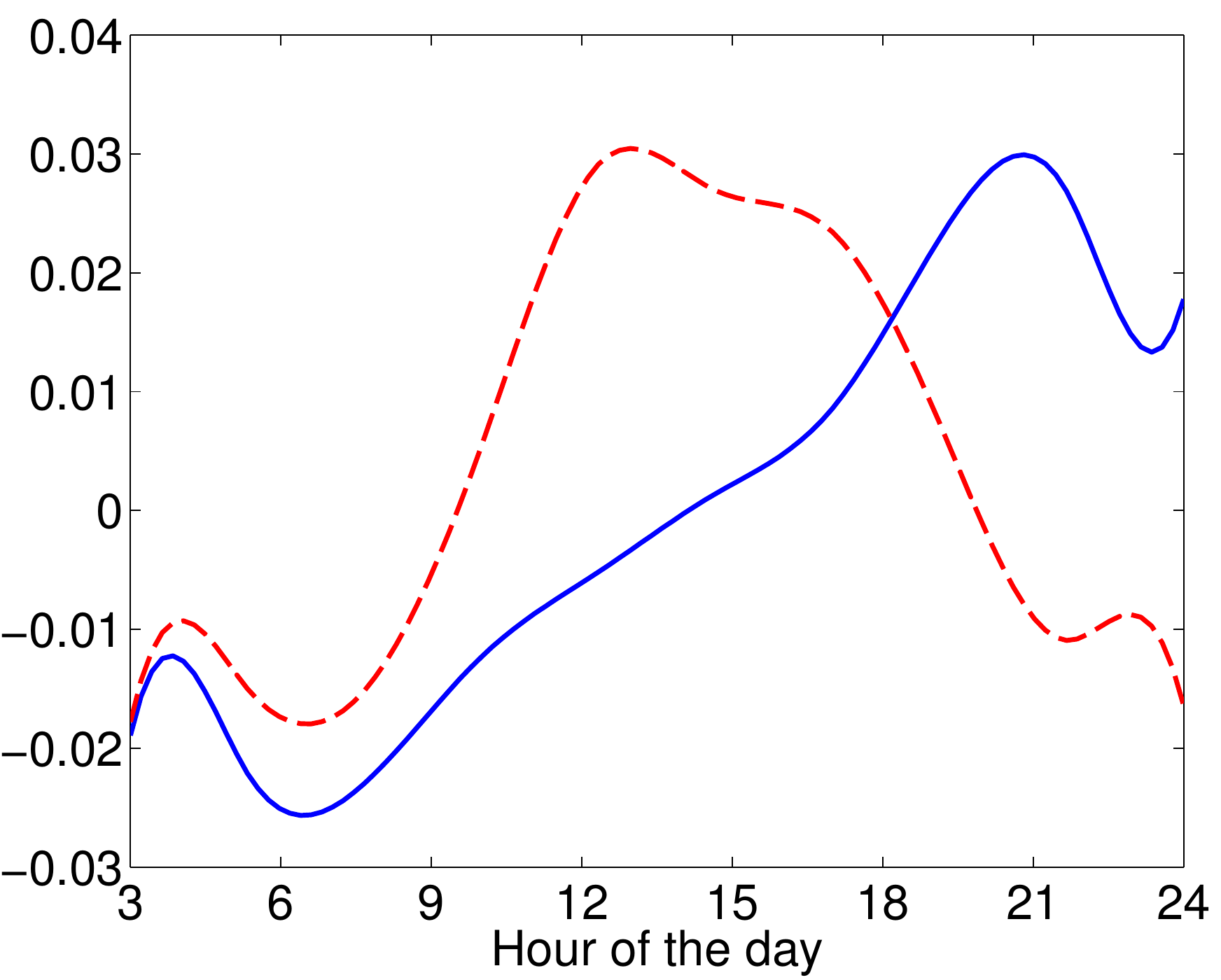}
}
\centering
\subfigure[]{
\includegraphics[width=0.2\textwidth]{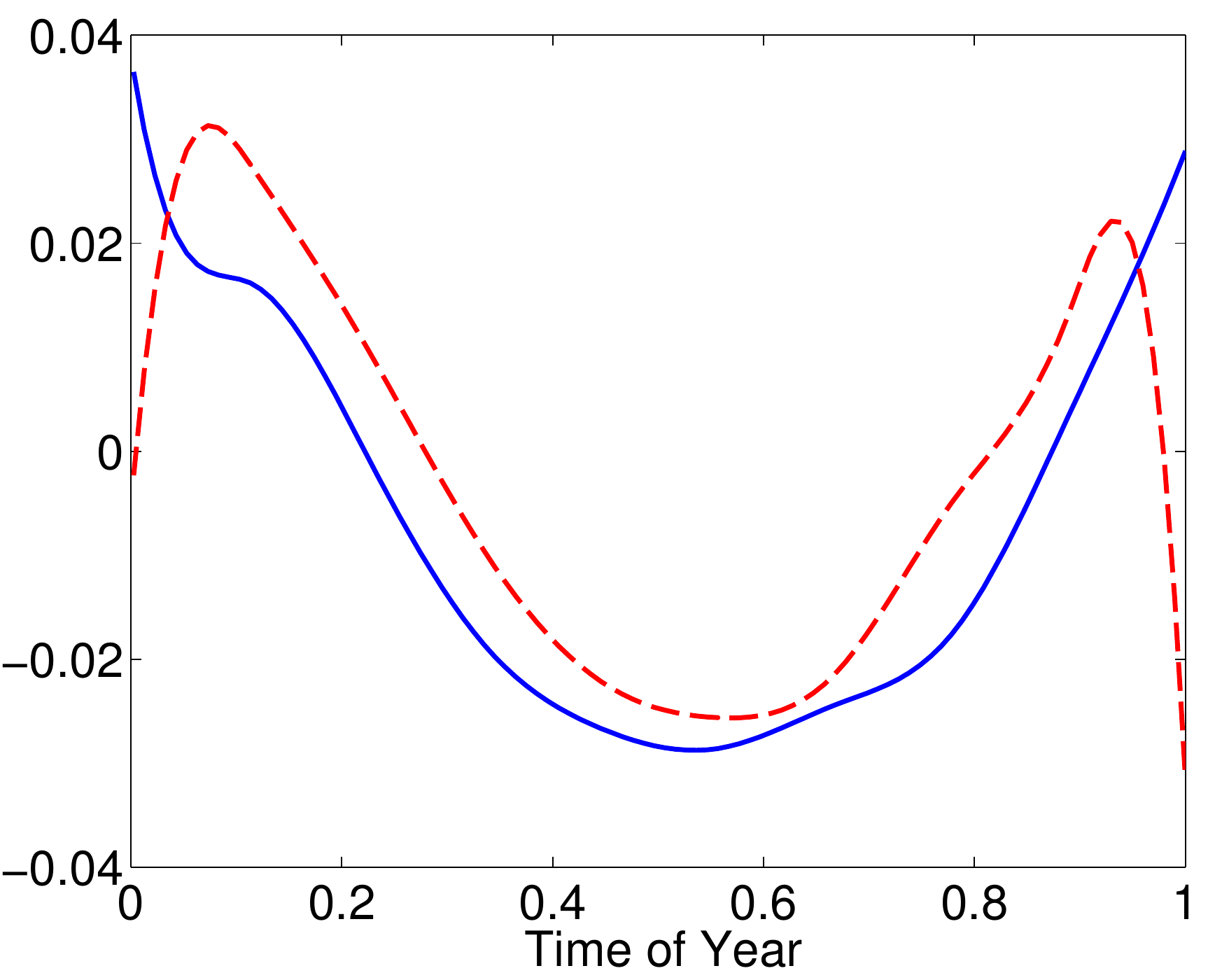}
}
\centering
\subfigure[]{
\includegraphics[width=0.2\textwidth]{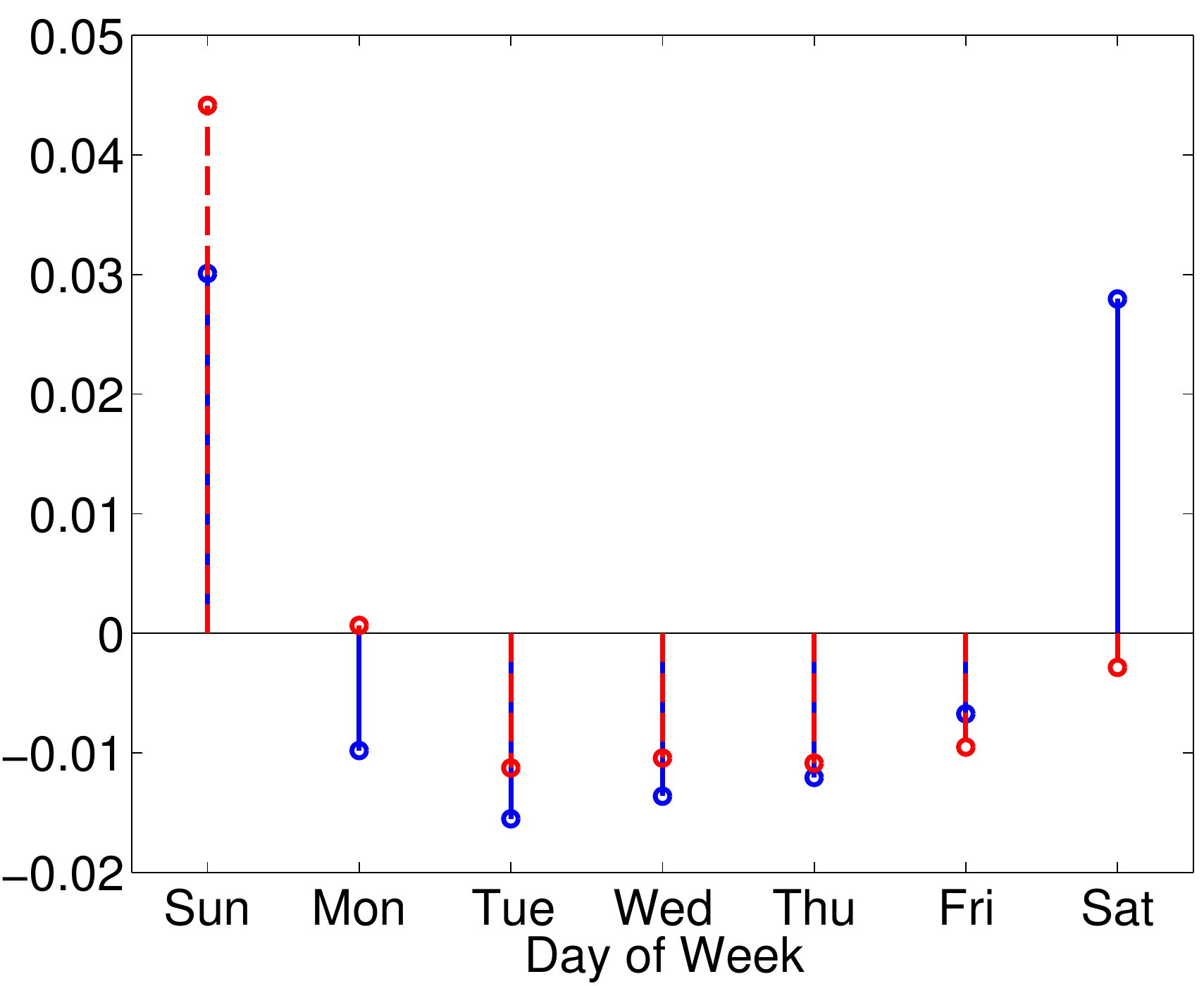}
}
\centering
\subfigure[]{
\includegraphics[width=0.2\textwidth]{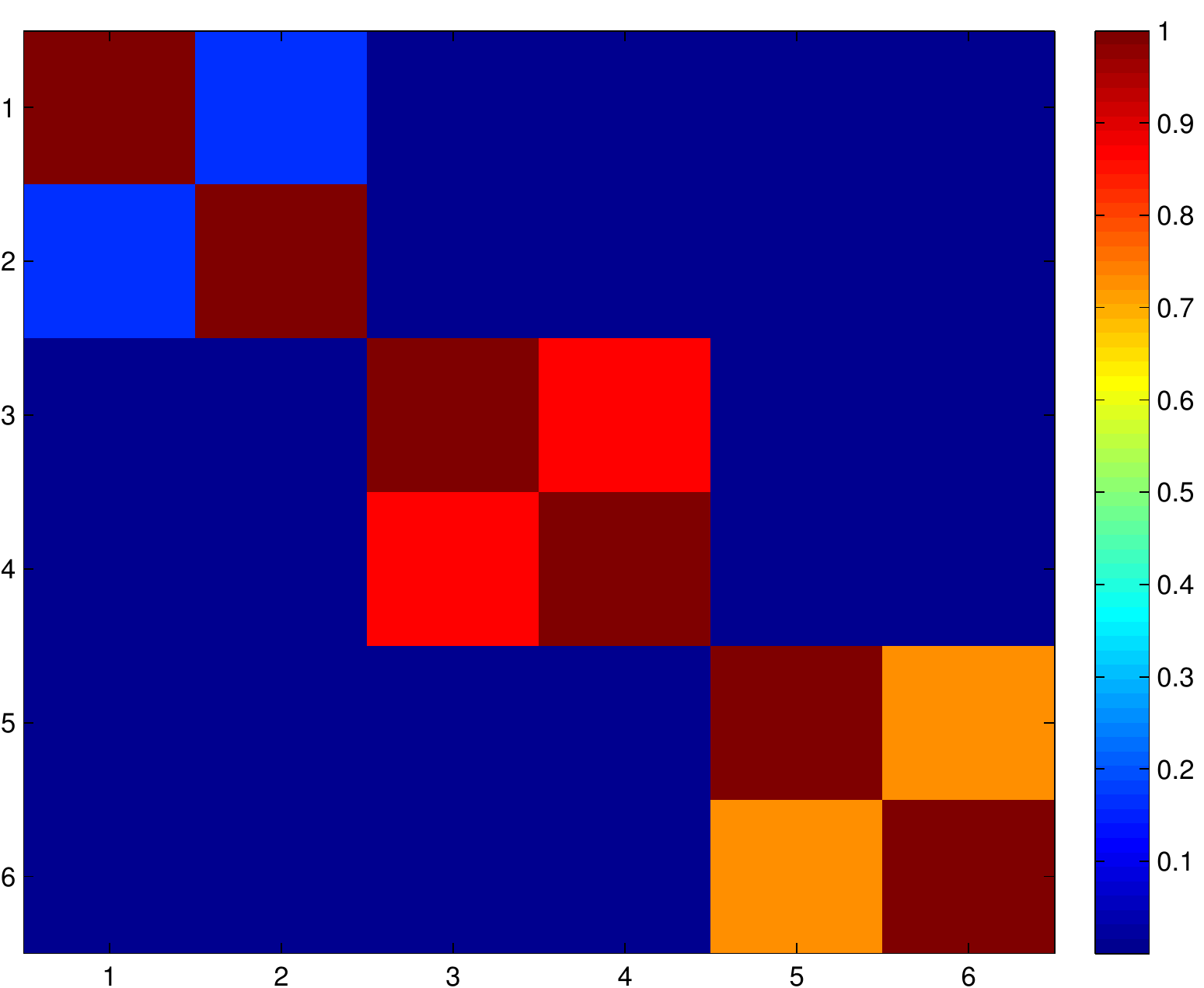}
}
\caption{\label{fig:tfs_L2} Transfer functions obtained with $L = 2$ (a-c), and correlation between atoms (d). (a): Hour of Day, (b): Time of Year (0: January, $1$st, 1: December, $31$st), (c): Day of Week.}
\end{figure*}

Fig.~\ref{fig:tfs_L2} (a)-(c) display the transfer functions obtained by our method for $L = 2$, and
Fig.~\ref{fig:tfs_L2} (d) shows the corresponding matrix of correlations $| \mbf{D}^T \mbf{D} |$ between the atoms of the dictionary estimated using the proposed algorithm\footnote{In this experiment, the observed covariates are equal for all the tasks (i.e., $x_{ij}^{(m)} = x_{ij}$), since Hour of day, Time of year and Day of week are clearly independent of the task at hand. This leads to a unique dictionary $\mathbf{D}$ that is independent of the task. See Sec. \ref{sec:multitask model formulation} for further details.}. As noted in Section \ref{sec:recovery_cond_COMP}, this matrix has a block diagonal structure due to similarites between transfer functions depending on the same covariate on the one hand, and independence of different covariates on the other hand. We obtain the coherence values $\mu_{intra}=0.87$ and $\mu_{inter}=0.01$ (see Definition \ref{def:our_coherence}), satisfying our recovery condition for BC-OMP in Theorem \ref{theorem:our_recovery_condition}, while the original condition in Theorem \ref{theorem:recovery_condition_OMP} for recovery in OMP is clearly not satisfied. 

% In the light of Section \ref{sec:}, C-OMP is guaranteed to rec

Let us consider the interpretability of the transfer functions learned using our method, and study correspondances with the
customer survey information in the CER data set.
Table \ref{tab:TF_residential_SME} relates the activation of the ``Hour of Day'' transfer functions to the customer type 
(residential vs.~SME). In this experiment we chose $L=2$, resulting in $4$ ``final'' transfer functions due to the non-negativity trick discussed in Sec. \ref{sec:weights_update}. 
One can see that, for residential customers, overwhelmingly the first transfer function is activated, while the majority of SME signals is modeled using the second one. Looking at the shape of the transfer functions, this intuitively makes sense:
the consumption of residential customers typically peaks in the evening, while SMEs consume most electricity during the day.
Similarly, Table \ref{tab:TF_SME_daytype} shows the correspondence between the activation of the ``Day of Week'' transfer function, and the SME business days (which is available from the CER survey information).
Again, there is an intuitive and easy-to-interpret correspondance between the learned models and available ground truth information.

\begin{table*}[t]
  \centering
   \caption{\label{tab:TF_residential_SME} Percentage of the activation of ``Hour of Day'' transfer functions for residential and SME customers.}
  \begin{tabular}{ |l||m{2.4cm} | m{2.4cm} | m{2.4cm} | m{2.4cm}|}
    \hline
      &
    \begin{minipage}{.5\textwidth}
     \vspace{1mm}
     \hspace{-3mm}
      \includegraphics[width=0.3\linewidth, height=20mm]{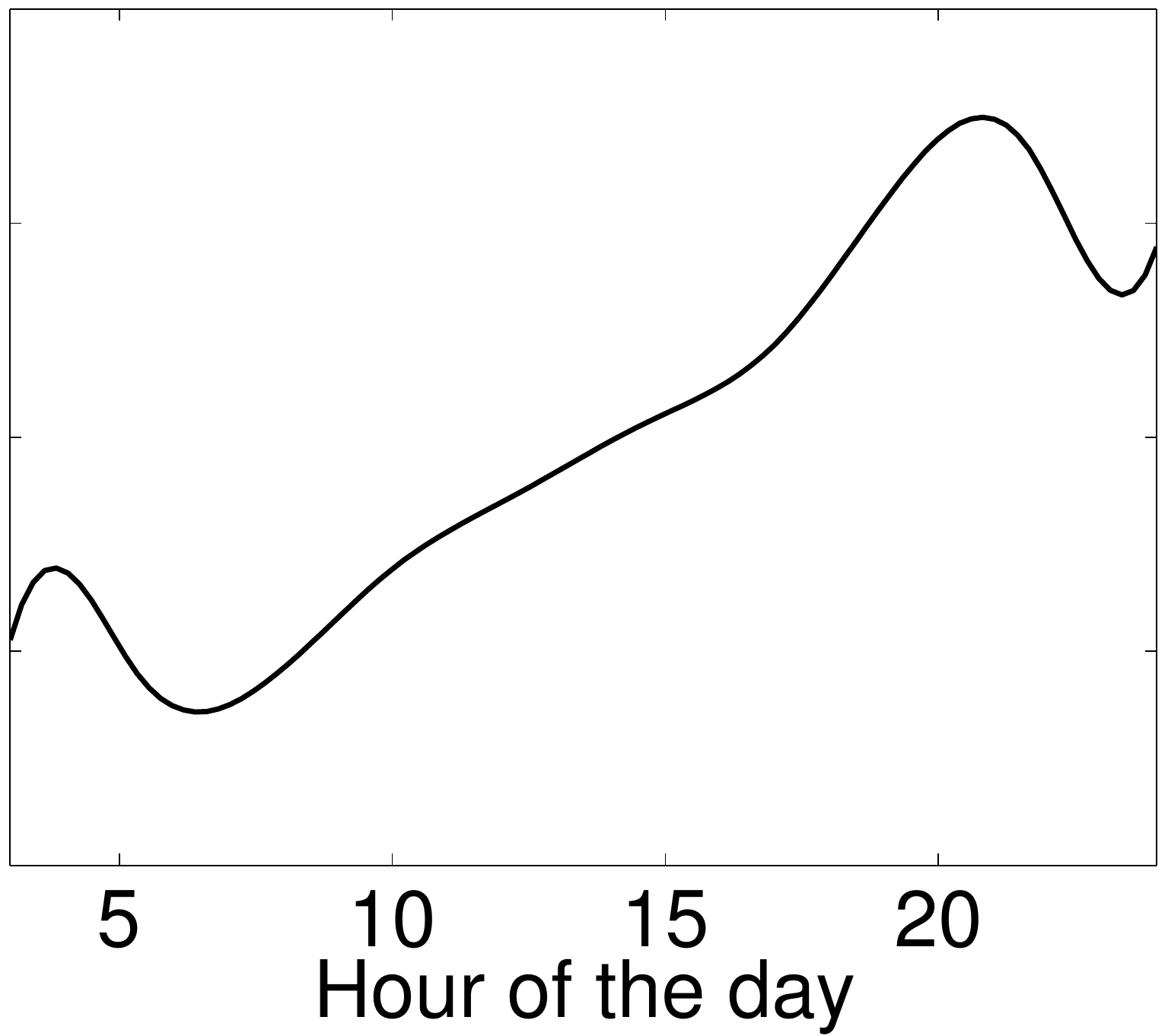}
     \vspace{1mm}
    \end{minipage}
    &
     \begin{minipage}{.5\textwidth}
     \vspace{1mm}
     \hspace{-3mm}
      \includegraphics[width=0.3\linewidth, height=20mm]{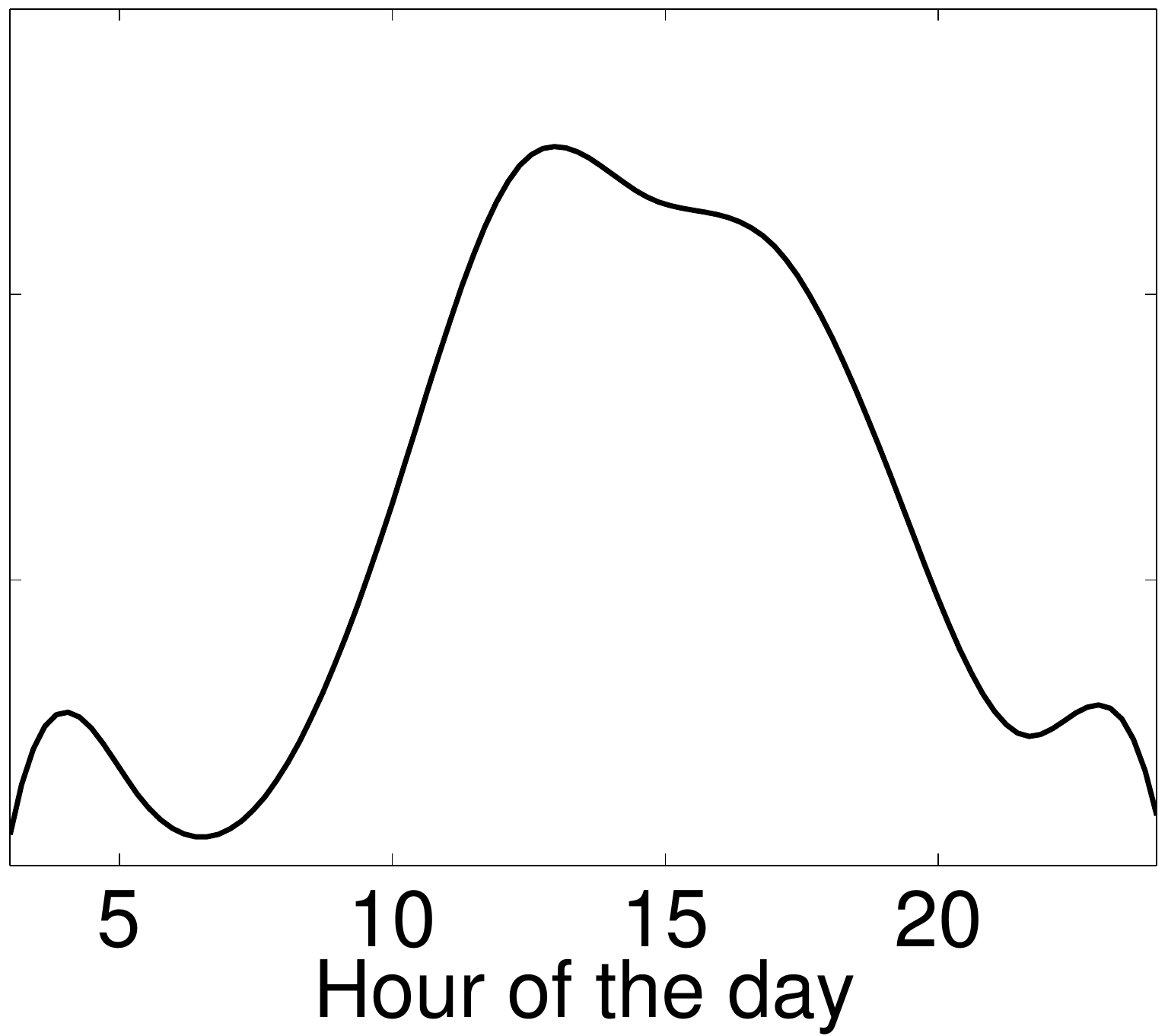}
      \vspace{1mm}
    \end{minipage}
    & 
     \begin{minipage}{.5\textwidth}
      \vspace{1mm}
	  \hspace{-3mm}      
      \includegraphics[width=0.3\linewidth, height=20mm]{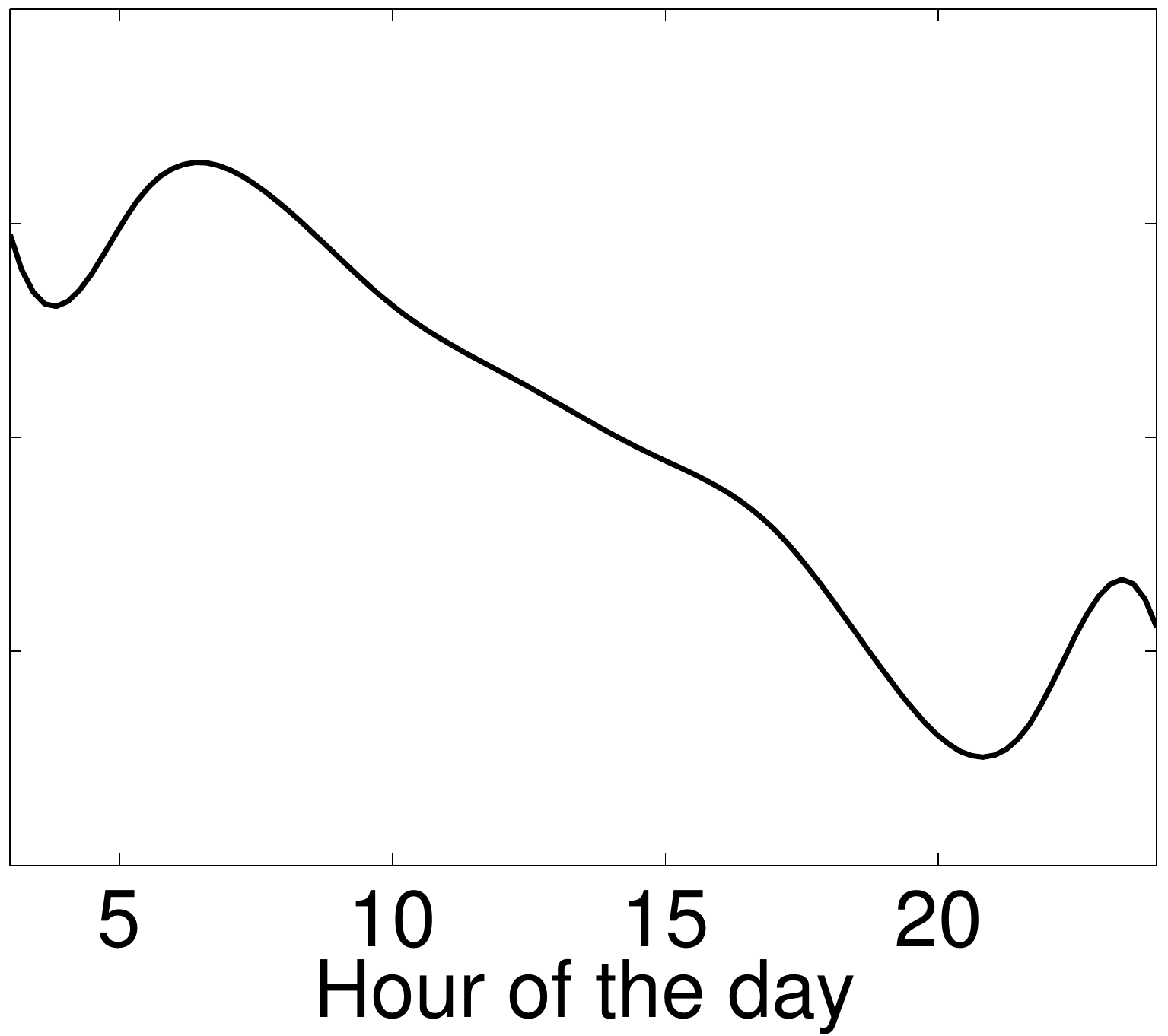}
      \vspace{1mm}
    \end{minipage}
    &
      \begin{minipage}{.5\textwidth}
      \vspace{1mm}
      \hspace{-3mm}
      \includegraphics[width=0.3\linewidth, height=20mm]{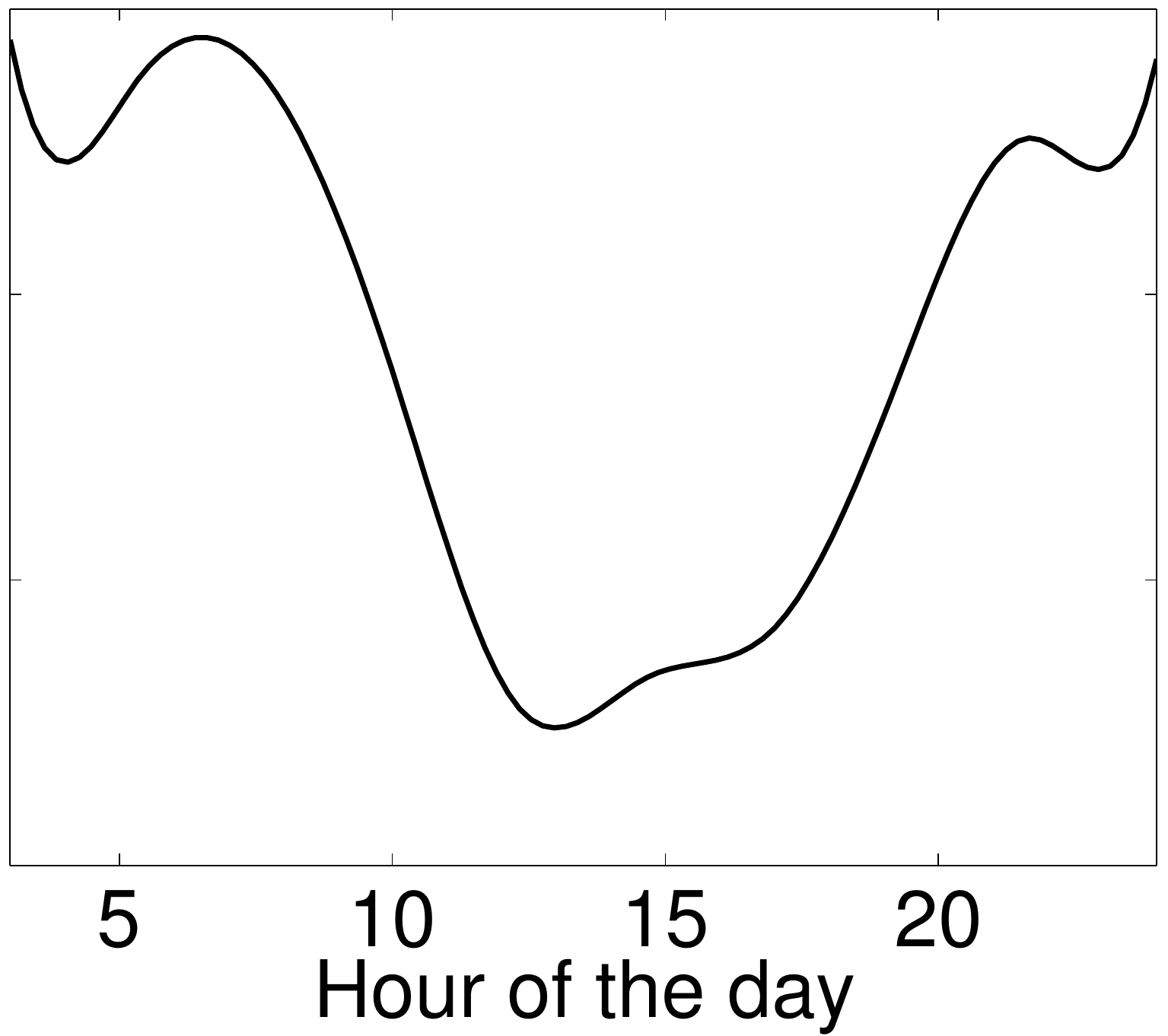}
      \vspace{1mm}
    \end{minipage}
    \\ \hline
    Residential & \textbf{89\%} & 9\% & 0\% & 2\% \\ \hline
    SME & 19\% & \textbf{68\%} & 8 \% & 5 \% \\ \hline
  \end{tabular}
\end{table*}

%\begin{table*}[ht]
%  \centering
%  \begin{tabular}{ | c | m{3cm} | m{3cm} | m{3cm} | m{3cm}|}
%    \hline
%      &    
%    \begin{minipage}{.3\textwidth}
%     \vspace{1mm}
%      \includegraphics[width=0.5\linewidth, height=20mm]{TF_TOD_1.pdf}
%     \vspace{1mm}
%    \end{minipage}
%    &
%     \begin{minipage}{.3\textwidth}
%     \vspace{1mm}
%      \includegraphics[width=0.5\linewidth, height=20mm]{TF_TOD_2.pdf}
%      \vspace{1mm}
%    \end{minipage}
%    & 
%     \begin{minipage}{.3\textwidth}
%      \vspace{1mm}
%      \includegraphics[width=0.5\linewidth, height=20mm]{TF_TOD_Neg1.pdf}
%      \vspace{1mm}
%    \end{minipage}
%    &
%      \begin{minipage}{.3\textwidth}
%      \vspace{1mm}
%      \includegraphics[width=0.5\linewidth, height=20mm]{TF_TOD_Neg2.pdf}
%      \vspace{1mm}
%    \end{minipage}
%    \\ \hline
%    Residential & \textbf{3250} & 325 & 8 & 56 \\ \hline
%    SME & 84 & \textbf{288} & 33 & 22 \\ \hline
%  \end{tabular}
%  \caption{\label{tab:TF_residential_SME} Percentage of the activation of transfer functions, for residential and SME households.}
%\end{table*}
% Original: width = 0.5, height = 20mm

\begin{table*}[t]
\caption{\label{tab:TF_SME_daytype} Percentage of the activation of ``Day of Week'' transfer functions for SMEs with different business days (see the left column).}
  \centering
  \begin{tabular}{|l|| m{2.4cm} | m{2.4cm} | m{2.4cm} | m{2.4cm}|}
    \hline
      &    
    \begin{minipage}{.3\textwidth}
     \vspace{1mm}
     \hspace{-3mm}
      \includegraphics[width=0.5\linewidth, height=20mm]{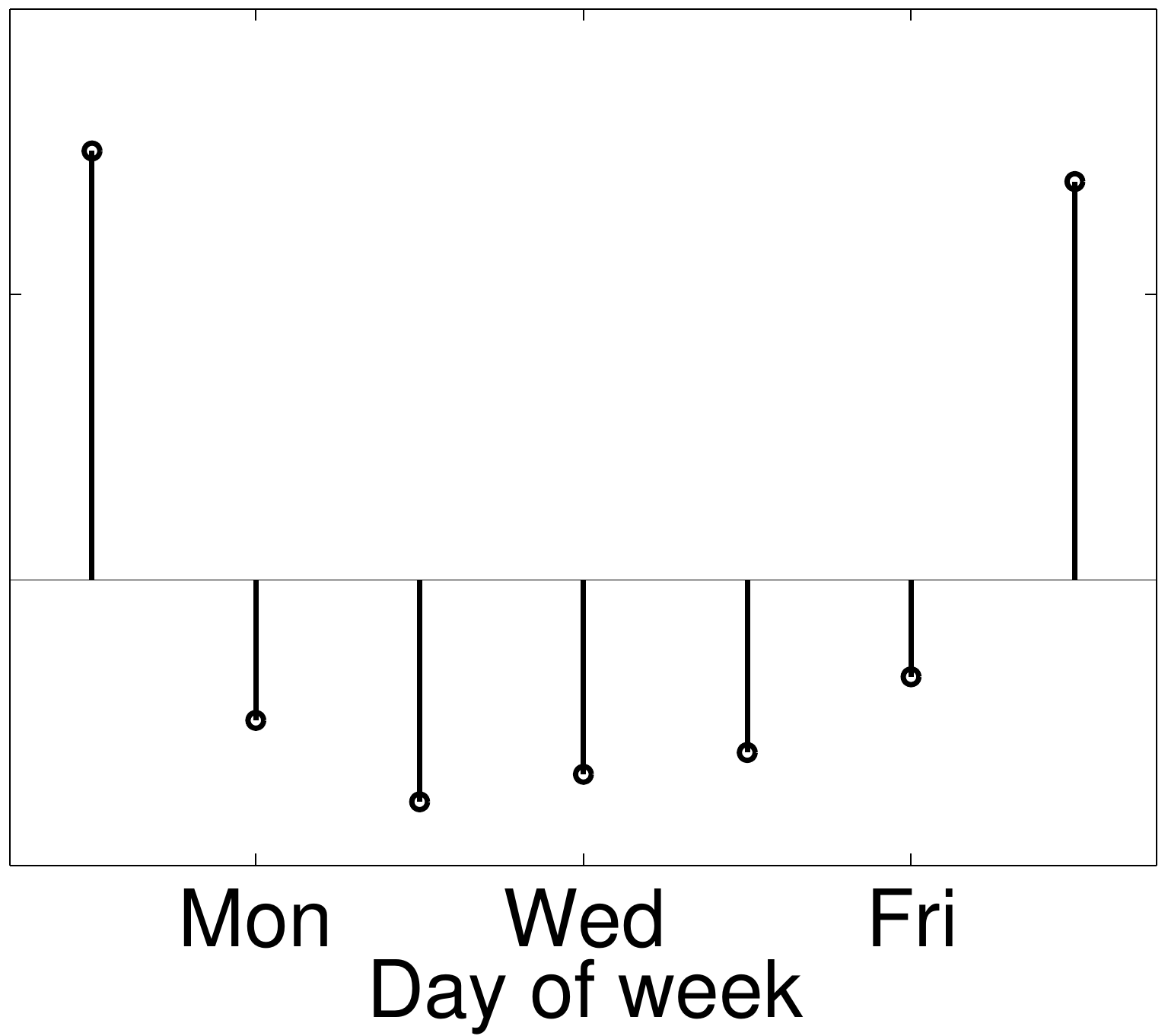}
     \vspace{1mm}
    \end{minipage}
    &
     \begin{minipage}{.3\textwidth}
     \vspace{1mm}
     \hspace{-3mm}
      \includegraphics[width=0.5\linewidth, height=20mm]{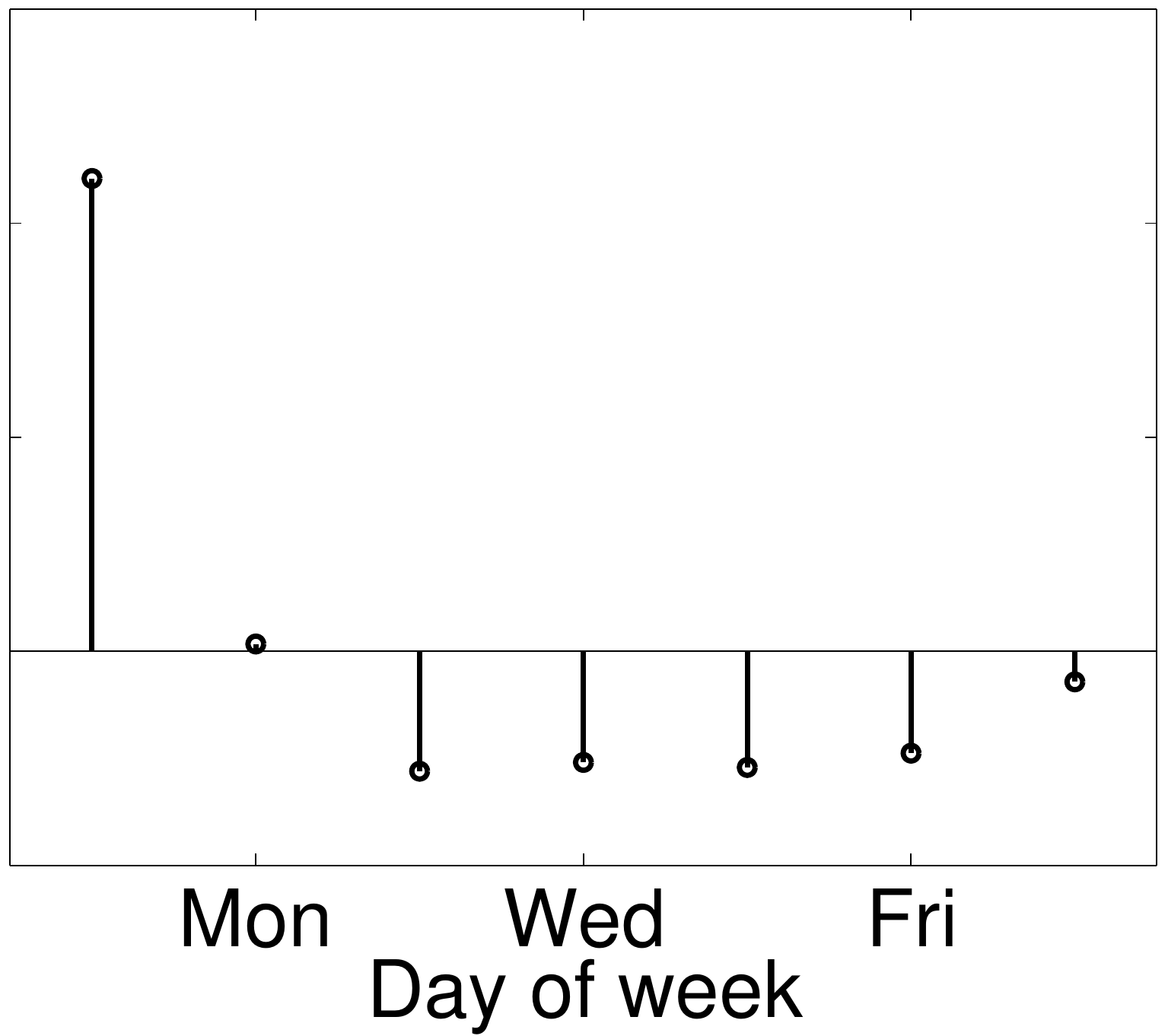}
      \vspace{1mm}
    \end{minipage}
    & 
     \begin{minipage}{.3\textwidth}
      \vspace{1mm}
      \hspace{-3mm}
      \includegraphics[width=0.5\linewidth, height=20mm]{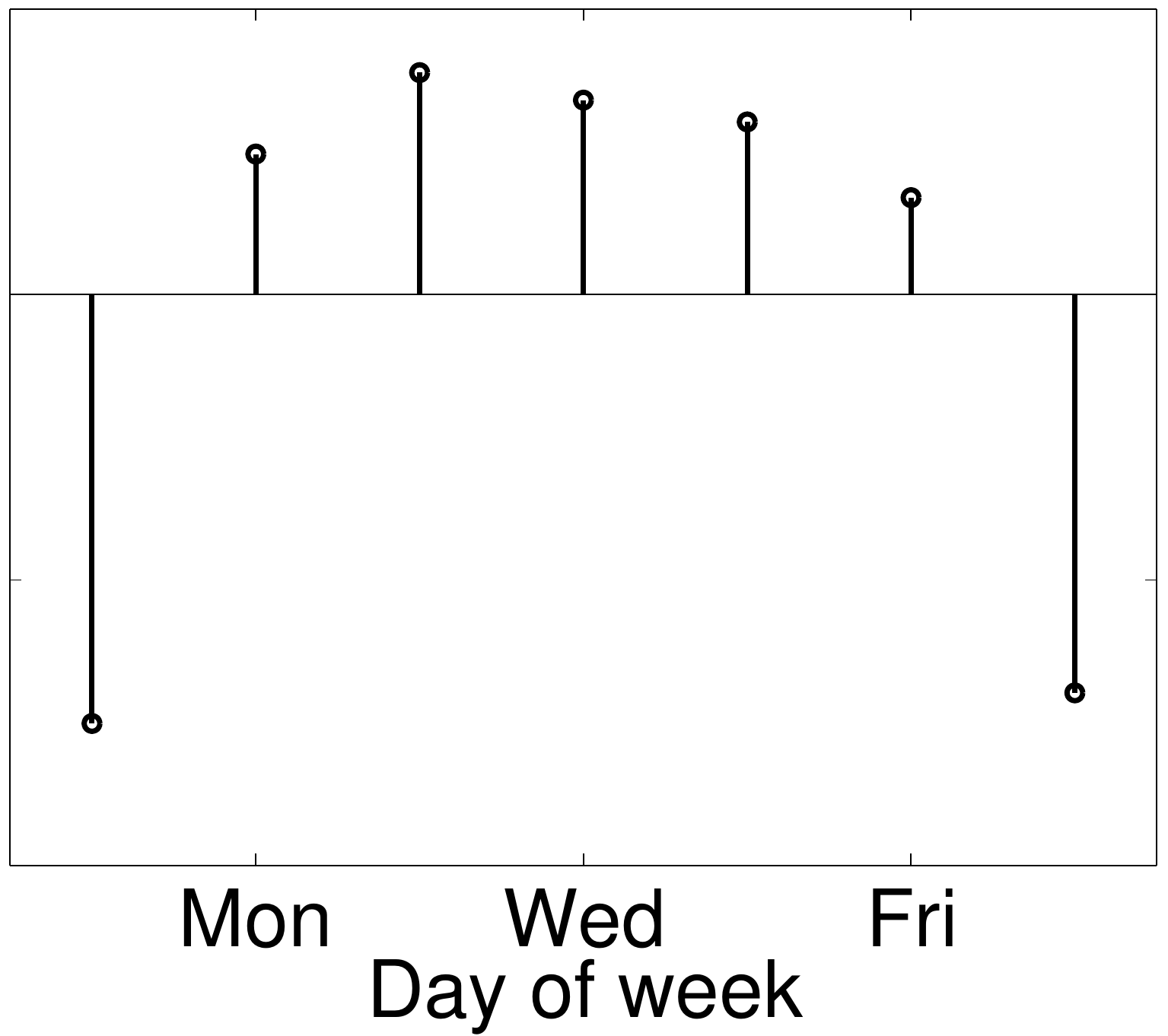}
      \vspace{1mm}
    \end{minipage}
    &
      \begin{minipage}{.3\textwidth}
      \vspace{1mm}
      \hspace{-3mm}
      \includegraphics[width=0.5\linewidth, height=20mm]{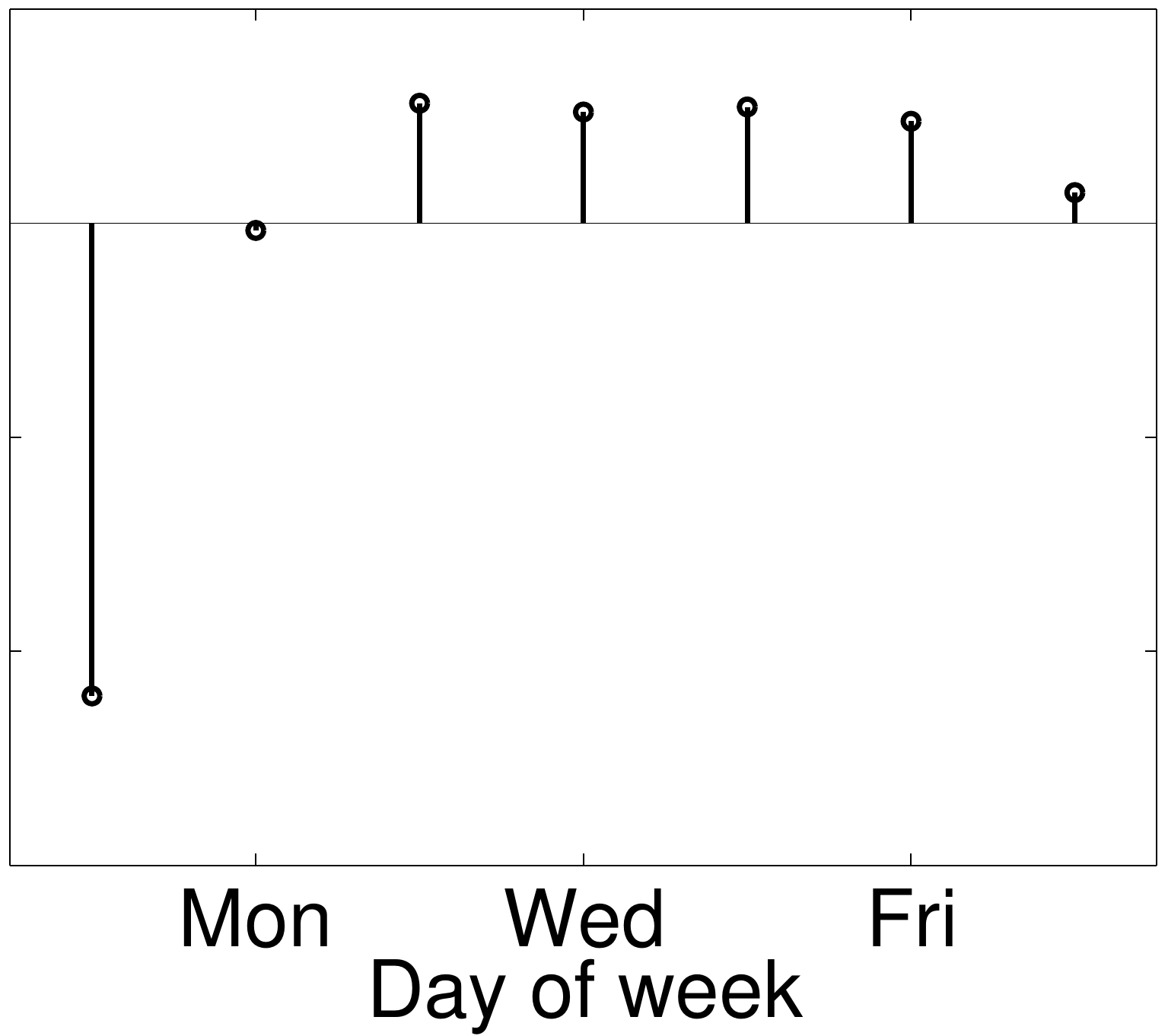}
      \vspace{1mm}
    \end{minipage}
    \\ \hline
    \footnotesize{Week days only} & 1\% & 2\% & \textbf{93\%} & 4\% \\ \hline
    \footnotesize{Week days + Saturday} & 0\% & 0\% & 37\% & \textbf{63\%} \\ \hline
    \footnotesize{All days} & 36\% & 5\% & 19\% & \textbf{40\%} \\ \hline
  \end{tabular}
\end{table*}

\begin{figure}[t]
\centering
\includegraphics[width=0.4\textwidth]{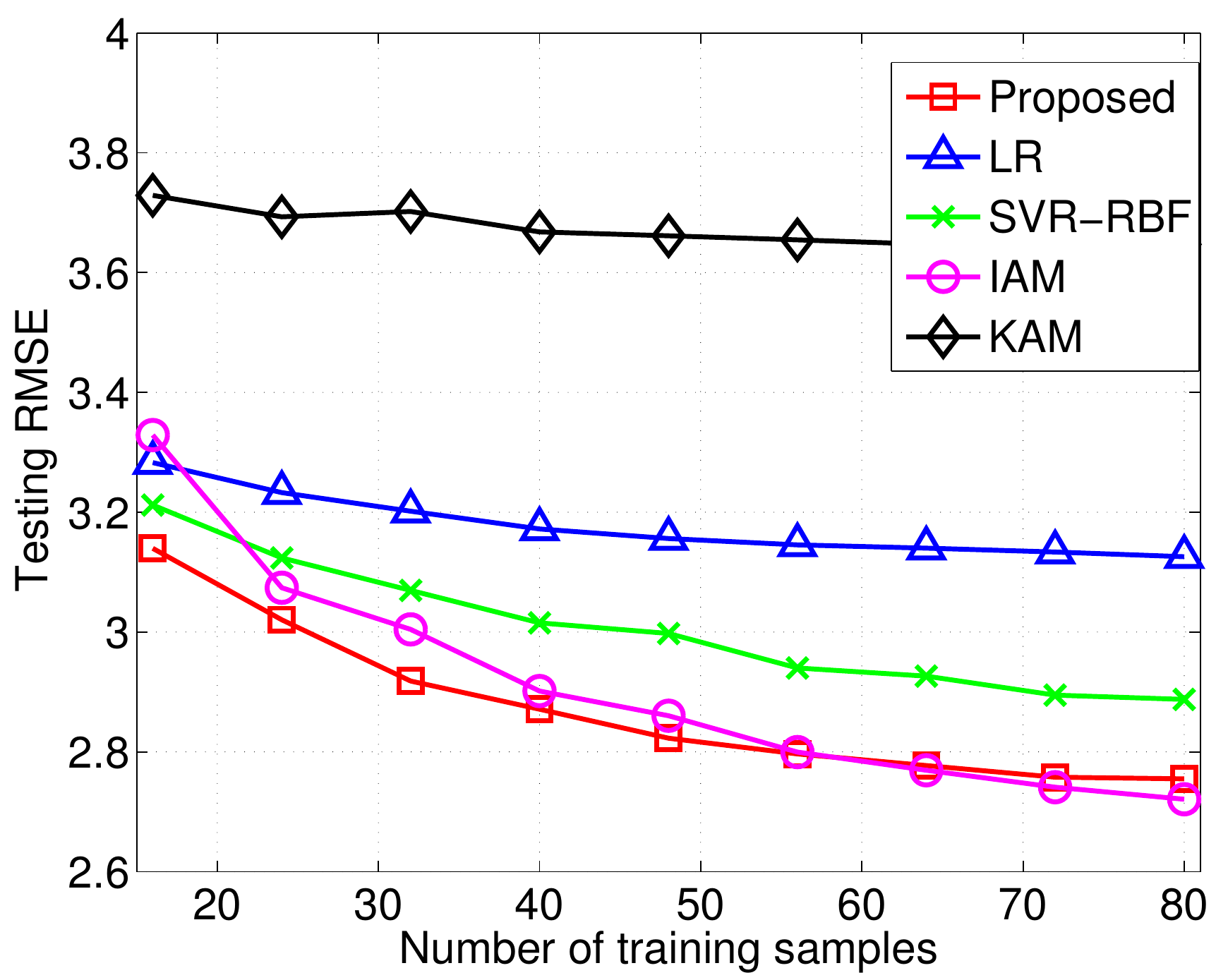}
\caption{\label{fig:testing_rmse_vs_numTrainingSamples} Testing RMSE versus number of training samples for the CER experiment.}
\end{figure}

Finally, we evaluate the performance of our method in a setting where training data is scarce. For this purpose, we consider now a training set of $n$ samples that are randomly selected from the CER data, and consider the remaining data for testing. Fig. \ref{fig:testing_rmse_vs_numTrainingSamples} illustrates the testing RMSE of the proposed method (with $L=2$) and the other competing methods with respect to $n$. It can be seen that when the training data is scarce, the proposed method outperforms IAM, and our method inherits the advantages of traditional multi-task learning by sharing information across tasks, and hence avoids overfitting. Note that for larger $n$, the gap between the two methods decreases, and IAM slightly outperforms our approach (with $L=2$), as it provides a much more flexible model, which however suffers from lack of interpretability. Note finally that our approach consistently outperforms all other competing methods (LR, SVR-RBF, KAM) in the range of training samples in Fig. \ref{fig:testing_rmse_vs_numTrainingSamples}.

% For this purpose, we split the CER data randomly into training and test tests, where the training set consists of only $1\%$ of the time points.
% Fig.~\ref{fig:rmse_vs_L_smallTraining} shows the results for IAM and our method with different values of $L$.
% Contrarily to the experiment behind Fig.~\ref{fig:rmse_vs_L}, smaller values of $L$ in this case yield better performance than IAM which suffers from overfitting. Hence, in settings where training data is lacking, our method inherits the advantages of traditional multi-task learning by sharing information across tasks.

%\begin{figure}[t!]
%\centering
%\includegraphics[width=0.2\textwidth]{mean_RMSE_vs_L_smallTraining.pdf}
%\caption{\label{fig:rmse_vs_L_smallTraining} Average RMSE on the CER data set vs.~parameter $L$ for our proposed method. The dashed line shows the results for IAM.}
%\end{figure}

\subsection{Intra-signal multi-task learning}
\label{sec:intra_signal_additive}
% \red{Add MAPE here?}
% We now turn to another application that is the prediction of 

In this last set of experiments, we consider another application of our multi-task learning framework.
A common approach in hourly electrical load forecasting is to treat each hourly period separately and use
different models for each hour of the day (see, e.g., \cite{fan2012short} and \cite{fay200324}).
Besides the computational burden, such approaches unfortunately fail to discover \textit{intra-daily} commonalities in the electricity consumption during different hours. Moreover, the resulting models are difficult to interpret.

We address those issues using the proposed multi-task framework. Given a signal $\mbf{y} \in \mathbb{R}^n$ representing hourly electrical loads, we first reshape the signal into a matrix $\mbf{Y} \in \mathbb{R}^{d \times 24}$, where  the $d$ rows represent the days, and the columns the hours in a day.
% This preprocessing step is illustrated in Fig. \ref{fig:preprocessing_step}. 
We then treat the columns of $\mbf{Y}$ as separate tasks, and fit our model using the proposed algorithm. 
% \red{CER or GEFCOM?}
%\begin{figure}[t!]
%\centering
%\includegraphics[width=0.51\textwidth]{image_reshape.pdf}
%\caption{\label{fig:preprocessing_step}Preprocessing step to create $24$ signals, out of a signal $\mbf{y}$.}
%\end{figure}
For this experiment, we use $4.5$ years of data from the GEFCom 2012 load forecasting challenge \cite{hong2014global}, considering ``Time Of Year'', ``Day of Week'' and ``Temperature'' as covariates. % Note that, unlike ``Time of Year'' and ``Day of Week'' covariates, the ``Temperature'' depends on the task (i.e., on the hour of the day).
The response variable is set to be the sum of the $20$ zonal level series expressed in gigawatt. Moreover, the temperature covariates are obtained by computing the average signals over the $11$ weather stations provided in the data. 
%\footnote{Unlike the CER experiment in Sec. 6.2 of the paper, the covariates considered here are quite correlated. Indeed, due to the seasonality, ``Time of Year'' and ``Temperature'' are quite correlated. This experiment shows nevertheless that the proposed method achieves good results in this scenario.}. 
%Unlike the previous experiment, this data is collected from the US where temperature plays a driving role in the load consumption. 
% We obtain the temperature covariates by computing the average over the $11$ weather stations provided in the data. 
% Moreover, for our multi-task approach, where one observation corresponds 
We use the first $d = 1,642$ days in the data set. The first $4$ years of the data are considered for training, and the remaining for testing. 
% We split the data into $4$ years for training and the remaining $6$ months for testing.
\iffalse

The original data consists in loads collected from $20$ zones in the US and temperatures from $11$ stations during $4.5$ years. 

For the sake of this experiment, we first average the temperatures from all stations to get a global temperature. A daily global temperature is then obtained by taking the mean temperature over every day. Moreover, we only consider the first $d = 1642$ days, where temperature data is given. We consider a global forecasting task where the response variable is the sum of the $20$ zonal level series expressed in gigawatt. The set is split into a training set of $4$ years, and the rest (half a year) for testing.
%\footnote{Note that this experimental setup differs from the one used in GEFCom competition.}.
\fi

\begin{table}
  \caption{\label{table:results_kaggle} Training and testing RMSE for the GEFCom task.} 
  \label{tab:Kaggle_datasetresults}
  \centering
  \begin{tabular}{@{}lrr@{}}
  \toprule
   Method & RMSE training & RMSE testing \\ \midrule
   Proposed ($L=4$) & \textbf{0.11} & \textbf{0.17} \\ 
   LR & 0.62 & 0.50 \\
   SVR-RBF & \textbf{0.11} & 0.18 \\
   Additive Model (AM) & 0.13 & 0.19 \\
   IAM & \textbf{0.11} & 0.18 \\
   \hline
  \end{tabular}
\end{table}

%\begin{figure}[ht]
%\centering
%\includegraphics[width=0.3\textwidth]{intraSignal_clustering_RMSE_test_vs_L.pdf}
%\caption{\label{fig:testingRMSE_vs_L} RMSE vs.~$L$ on the GEFCom data, for the proposed method (solid) and IAM (dashed).}
%\end{figure}

We compare the proposed approach to single-task regression methods that do not split the signal into hourly signals. Specifically, considering an additional ``Hour of Day'' covariate, we fit linear and nonlinear models using LR, SVR-RBF as well as an Additive Model that reads
\begin{align*}
y_i & = f_1(\text{Hour of Day}_i) + f_2(\text{Time of Year}_i) \\ 
     & + f_3(\text{Day of Week}_i) + f_4(\text{Temperature}_i).
\end{align*}
In addition, we compare the proposed approach to IAM using the same split into hourly signals.

\begin{figure}[t]
\centering
\subfigure[]{
\includegraphics[width=0.2\textwidth]{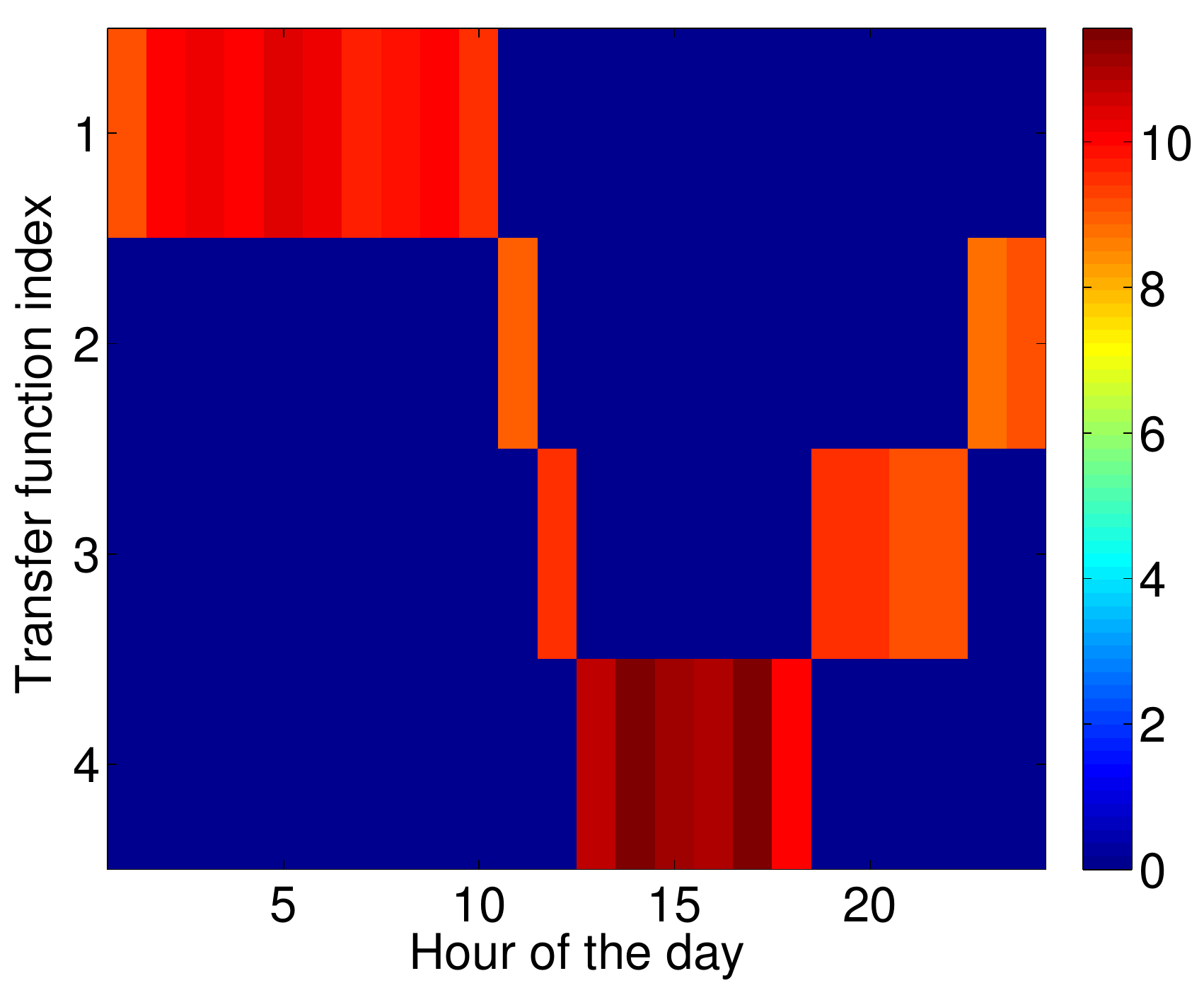}
}
\centering
\subfigure[]{
\includegraphics[width=0.2\textwidth]{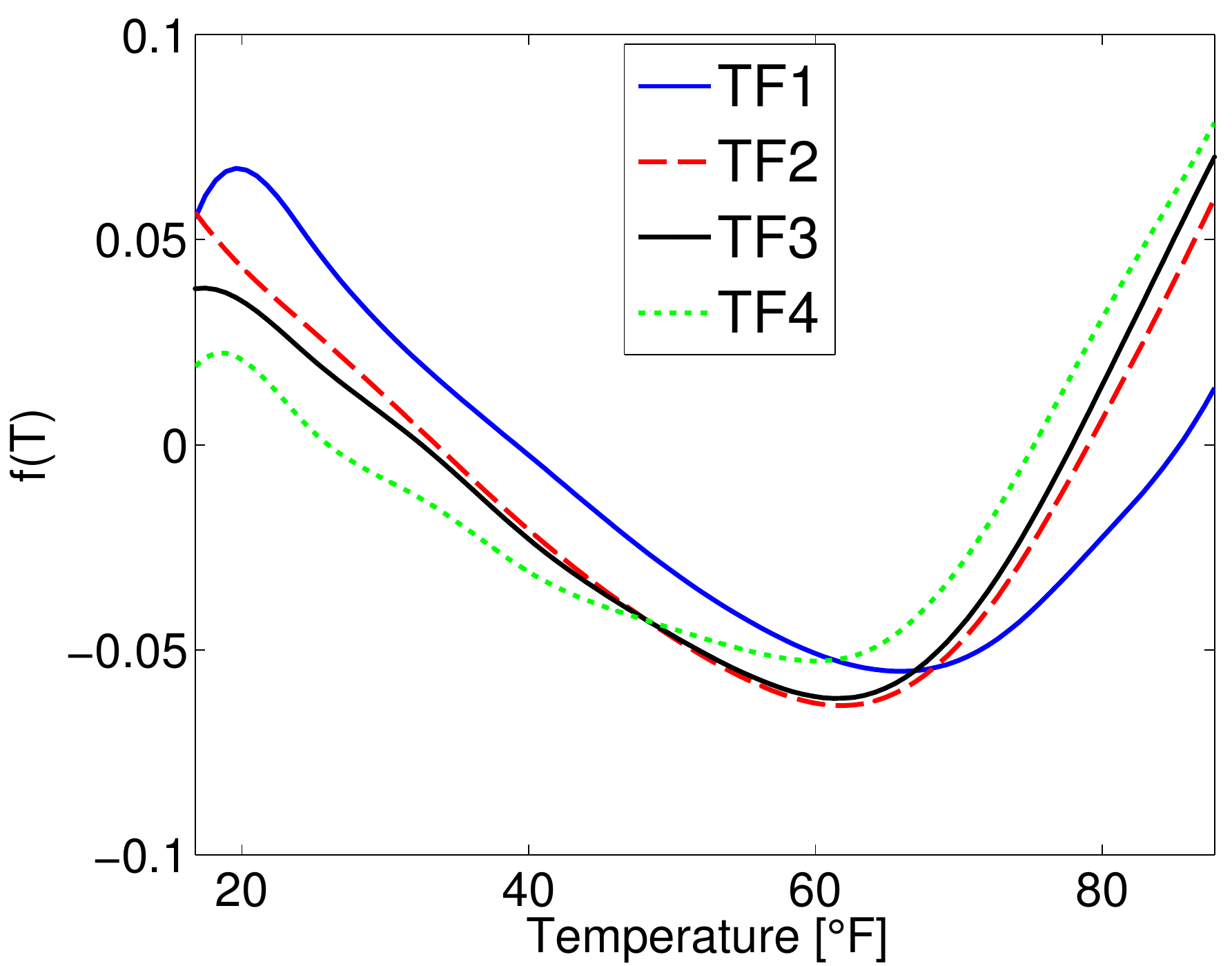}
}
\caption{\label{fig:intra_Q_TF} Results on the GEFCom data with $L = 4$. (a) $\boldsymbol\Lambda_{\text{temp}}$ matrix showing the activation of the temperature transfer functions for the 24 hours per day. (b) Shapes of the estimated temperature transfer functions.}
\end{figure}

Table \ref{table:results_kaggle} shows the result of the comparison. It can be seen that, with $L = 4$, the proposed approach outperforms all other methods in terms of testing RMSE. By splitting the signal into hourly signals, our algorithm yields a testing RMSE that has improved roughly by $10 \%$ with respect to AM. In addition, our approach slightly outperforms IAM in terms of testing accuracy in this experiment, while learning much less transfer functions. Interestingly, our algorithm yields a clustering of the hours of the day with some intuitive interpretations:
% Note that, even for small values, our method is competitive with IAM. In addition, it yields a clustering of the hours of the day with some intuitive interpretations:
consider the matrix $\boldsymbol\Lambda_{\text{temp}}$ in Fig.~\ref{fig:intra_Q_TF} (a) which shows the assignment of the $L=4$ temperature transfer functions (displayed in (b)) to the $24$ signals representing different hours per day. To be more specific, the matrix $\boldsymbol\Lambda_{\text{temp}}$ is given by
\[
\boldsymbol\Lambda_{\text{temp}} = 
\begin{bmatrix}
 \lambda_{j1}^{(1)} & \dots & \lambda_{j1}^{(24)} \\
\vdots & \vdots & \vdots \\
\lambda_{j4}^{(1)} & \dots & \lambda_{j4}^{(24)}
 \end{bmatrix}
 ,
\]
where $j$ corresponds to the index of the ``Temperature'' covariate. Note that the transitions are ``smooth'', i.e., consecutive hours are typically modeled using the same temperature transfer functions, albeit we did not explicitely enforce this property. 
While all the transfer functions in Fig.~\ref{fig:intra_Q_TF} (b) have a similar V-shape, there are noteworthy differences. 
For example, TF$4$ compared to TF$1$ leads to higher load predictions for hot temperatures and lower load predictions for cold temperatures. Intuitively, we can interpret TF$4$ and TF$1$ as representing ``air conditioning'' and ``heating'' effects, respectively. This corresponds well with the hours for which these two functions are activated: TF$4$ during the day where most air conditioning occurs, and TF$1$ during the night and early morning, where most electricity is used for heating.

\iffalse
% at large temperatures, the value of TF$i$ increases with $i$ (\red{right arrow}). Conversely, for small temperatures, TF$i$ is \textit{decreasing} with $i$ (\red{left arrow}). 
In light of Fig. \ref{fig:intra_Q_TF} (a), this makes sense as TF1 is active in the morning, where heating is usually ON and AC is OFF. On the other hand, AC is usually ON in the afternoon, when TF4 is active. 
% TF4 predicts high load values for large temperatures due to AC, which tends to be active in the afternoon. 
TF2 and TF3 represent a gradual transition, in time and in the values of the transfer functions, between the transfer functions TF1 and TF4.
\fi

%  \red{Interesting to note is also that ``sequential'' (?) nature of the assignments. This is a bit hard to explain...} 

% while the morning transfer function (TF1) predicts high 

% associated to the ``Temperature'' covariate, that shows the activation v
% and the temperature transfer functions in Fig. \ref{fig:intra_Q_TF} (b). 

% We illustrate the testing RMSE with respect to $L$ 

\vspace{-1mm}
\section{Conclusion}\label{sec:conclusions}

In this paper, we introduced a novel multi-task learning framework for additive models with the key idea
to \textit{share transfer functions across the different tasks}.
We established a connection between the proposed model and sparse dictionary learning and leveraged it to derive an efficient fitting algorithm. We further conducted a theoretical analysis of the recovery conditions of the sparse representation step; by distinguishing between coherence within and across different subdictionaries, we were able to establish recovery for a wider range of realistic settings that are particularly relevant in our multi-task learning problem. Through synthetic experiments, we showed that the proposed algorithm correctly estimates the underlying transfer functions, and outperforms competing methods in terms of predictive power. In experiments with real-world electricity demand data, we demonstrated that our proposed multi-task approach achieves competitive performance with baseline methods that learn models independently for each task, while providing models that are \textit{more interpretable}, extracting \textit{inherent structure} in the tasks (e.g., clustering of tasks corresponding to different customer types), and being \textit{more robust} in settings where training data are scarce.
In future work, we plan to improve the scalability of the method to apply it to domains that potentially involve millions of tasks.
\appendix

\subsection{Proof of Theorem \ref{theorem:our_recovery_condition}}

%\begin{theorem}[Recovery condition]
%\label{theorem:our_recovery_condition}
%If the following condition holds:
%\begin{align*}
%\mu_{intra} + 2 (p-1) \mu_{inter} < 1,
%\end{align*}
%then C-OMP recovers the correct atoms and their coefficients.
%\end{theorem}
% \begin{proof}
Assume that $\mbf{y} = \sum_{j=1}^p \gamma_j \mbf{d}_{j, l_j}$.
We prove by induction that the correct atoms $\mbf{d}_{j, l_j}$ are recovered when the sufficient condition holds. Assume that, after $j \in \{0, \dots, p-1\}$ steps, BC-OMP has recovered correct atoms in the support. Therefore, it holds that the residual signal $\mbf{r}_j \in \text{span} (\mbf{d}_{1, l_1}, \dots, \mbf{d}_{p, l_p})$ and we write
\begin{align*}
\mbf{r}_j = \sum_{g=1}^p \alpha_g \mbf{d}_{g, l_g}.
\end{align*}
Since $\mbf{y}$ is exactly $p$-sparse, the residual is non-zero, and $\boldsymbol\alpha \neq \mathbf{0}$.
The atom selected by BC-OMP at step $j+1$ is optimal if and only if:
\begin{align}
\label{eq:exact_condition}
\max_{k \in \mathcal{A}_j} \max_{\substack{l \in [L_k] \\ l \neq l_k}} \left| \scalprod{\mbf{r}_j}{\mbf{d}_{k, l}}\right| < \max_{k \in \mathcal{A}_j} \left| \scalprod{\mbf{r}_j}{\mbf{d}_{k, l_k}} \right|.
\end{align}
We establish the recovery condition by showing a lower bound to the right hand side and an upper bound to the left hand side of Eq. (\ref{eq:exact_condition}). Note that, for any $k \in [p]$:
\begin{align*}
\left| \scalprod{\mbf{r}_j}{\mbf{d}_{k, l_k}} \right| & = \left| \sum_{g=1}^p \alpha_g \scalprod{\mbf{d}_{g, l_g}}{\mbf{d}_{k, l_k}} \right| \\ & = \left| \alpha_k + \sum_{g \neq k} \alpha_g \scalprod{\mbf{d}_{g, l_g}}{\mbf{d}_{k, l_k}} \right| \\ & \geq |\alpha_k| - \sum_{g \neq k} |\alpha_g| \left|\scalprod{\mbf{d}_{g, l_g}}{\mbf{d}_{k, l_k}} \right|.
\end{align*}
Moreover, by definition, $\left| \scalprod{\mbf{d}_{g, l_g}}{\mbf{d}_{k, l_k}} \right| \leq \mu_{inter}$ for $g \neq k$. It follows that the right hand side of Eq. (\ref{eq:exact_condition}) can be bounded as follows:
\begin{align*}
\max_{k \in \mathcal{A}_j} \left| \scalprod{\mbf{r}_j}{\mbf{d}_{k, l_k}} \right| & \overset{(a)}{=} \max_{k \in [p]} \left| \scalprod{\mbf{r}_j}{\mbf{d}_{k, l_k}} \right| \\ & \geq \max_{k \in [p]} \left\{ |\alpha_k| - \mu_{inter} \sum_{g \neq k} |\alpha_g| \right\}  \\
& \overset{(b)}{\geq} \| \boldsymbol\alpha \|_{\infty} - (p-1) \mu_{inter}  \| \boldsymbol\alpha \|_{\infty},
\end{align*}
where (a) is due to the fact that atoms $\mbf{d}_{k, {l_k}}$ that are \textit{not} in $\mathcal{A}_j$ have already been selected, and are therefore orthogonal to $\mbf{r}_j$. Inequality (b) is obtained by bounding each term $|\alpha_g|$ by $\| \boldsymbol\alpha \|_{\infty}$.

We now exhibit an upper bound to the left hand side term of Eq. (\ref{eq:exact_condition}). We have:
\begin{align*}
& \max_{k \in \mathcal{A}_j} \max_{\substack{l \in [L_k]] \\ l \neq l_k}} \left| \scalprod{\mbf{r}_j}{\mbf{d}_{k,l}}\right| \\
\leq &  \max_{k \in [p]} \max_{\substack{l \in [L_k] \\ l \neq l_k}} \left| \scalprod{\mbf{r}_j}{\mbf{d}_{k,l}}\right| \\ 
= & \max_{k \in [p]} \max_{\substack{l \in [L_k] \\ l \neq l_k}} \left| \alpha_k \scalprod{\mbf{d}_{k, l_k}}{\mbf{d}_{k,l}} + \sum_{g \neq k} \alpha_g \scalprod{\mbf{d}_{g, l_g}}{\mbf{d}_{k,l}} \right| \\
																  \leq & \max_{k \in [p]} \left| \alpha_k \right| \mu_{intra} + \mu_{inter} \sum_{g \neq k} |\alpha_g| \\
																  \leq &  \mu_{intra} \| \boldsymbol\alpha \|_{\infty} + (p-1) \mu_{inter} \| \boldsymbol\alpha \|_{\infty}.
\end{align*}
Therefore, we obtain the following condition for Eq. (\ref{eq:exact_condition}) to hold:
\begin{align*}
\mu_{intra} \| \boldsymbol \alpha \|_{\infty} & + (p-1) \mu_{inter} \| \boldsymbol \alpha \|_{\infty} \\
																			& <  \| \boldsymbol\alpha \|_{\infty} - (p-1) \mu_{inter}  \| \boldsymbol\alpha \|_{\infty}.
\end{align*}
Since $\boldsymbol \alpha \neq \mbf{0}$, we simplify the condition to:
\begin{align*}
\mu_{intra} + 2 (p-1) \mu_{inter} < 1.
\end{align*}
As the above condition holds by assumption, we conclude that Eq. (\ref{eq:exact_condition}) is satisfied and BC-OMP selects a correct atom at step $j+1$.
% Once the correct support is recovered, it is straightforward to see that the correct coefficients are also correctly recovered by orthogonal projection into the atoms in the support. This concludes the proof.

\noindent Once the correct support is recovered, it is straightforward to see that an orthogonal projection onto the span of the recovered atoms yields the correct coefficients. Indeed, if we have $\mbf{y} = \sum_{j=1}^p \gamma_j' \mbf{d}_{j, l_j}$, the linear independence of the atoms $\{ \mbf{d}_{j, l_j} \}$ imposes $ \gamma_j' = \gamma_j$. This concludes the proof. $\qed$

\bibliographystyle{IEEEbib}
\bibliography{reference}

\begin{thebibliography}{10}

\bibitem{hastie2009elements}
T~Hastie, R~Tibshirani, and J~Friedman,
\newblock {\em The elements of statistical learning},
\newblock Springer, 2009.

\bibitem{wood2006generalized}
S~Wood,
\newblock {\em Generalized additive models: an introduction with R},
\newblock CRC press, 2006.

\bibitem{hastie1990generalized}
T~Hastie and R~Tibshirani,
\newblock {\em Generalized additive models}, vol.~43,
\newblock CRC Press, 1990.

\bibitem{aharon2006svd}
M~Aharon, M~Elad, and A~Bruckstein,
\newblock ``K-svd: An algorithm for designing overcomplete dictionaries for
  sparse representation,''
\newblock {\em IEEE Trans. on Sig. Proc.}, vol. 54, no. 11, pp. 4311--4322,
  2006.

\bibitem{tosic2011dictionary}
I~Tosic and P~Frossard,
\newblock ``Dictionary learning,''
\newblock {\em IEEE Sig. Proc. Mag.}, vol. 28, no. 2, pp. 27--38, 2011.

\bibitem{kreutz2003dictionary}
K~Kreutz-Delgado, J~Murray, B~Rao, K~Engan, TW~Lee, and T~Sejnowski,
\newblock ``Dictionary learning algorithms for sparse representation,''
\newblock {\em Neural computation}, vol. 15, no. 2, pp. 349--396, 2003.

\bibitem{pati1993orthogonal}
YC~Pati, R~Rezaiifar, and PS~Krishnaprasad,
\newblock ``Orthogonal matching pursuit: Recursive function approximation with
  applications to wavelet decomposition,''
\newblock in {\em Asilomar Conference on Signals, Systems and Computers}, 1993.

\bibitem{Mallat93}
S~Mallat and Z~Zhang,
\newblock ``Matching pursuits with time-frequency dictionaries,''
\newblock {\em IEEE Trans. on Sig. Proc.}, vol. 41, pp. 3397--3415, Dec. 1993.

\bibitem{engan1999method}
K~Engan, S~Aase, and Hakon H,
\newblock ``Method of optimal directions for frame design,''
\newblock in {\em IEEE ICASSP}, 1999, vol.~5, pp. 2443--2446.

\bibitem{caruana1997multitask}
R~Caruana,
\newblock ``Multitask learning,''
\newblock {\em Machine Learning}, vol. 28, pp. 41--75, 1997.

\bibitem{bakker2003task}
B~Bakker and T~Heskes,
\newblock ``Task clustering and gating for bayesian multitask learning,''
\newblock {\em JMLR}, vol. 4, pp. 83--99, 2003.

\bibitem{Thrun_1997_622}
S~Thrun and L~Pratt,
\newblock {\em Learning To Learn},
\newblock Kluwer Academic Publishers, November 1997.

\bibitem{evgeniou2005learning}
T~Evgeniou, C~Micchelli, and M~Pontil,
\newblock ``Learning multiple tasks with kernel methods,''
\newblock in {\em JMLR}, 2005, pp. 615--637.

\bibitem{evgeniou2007multi}
A~Evgeniou and M~Pontil,
\newblock ``Multi-task feature learning,''
\newblock {\em NIPS}, vol. 19, pp. 41--48, 2007.

\bibitem{jacob2009clustered}
L~Jacob, J-P Vert, and F~Bach,
\newblock ``Clustered multi-task learning: A convex formulation,''
\newblock in {\em NIPS}, 2009, pp. 745--752.

\bibitem{liu2009nonparametric}
H~Liu, L~Wasserman, and J~Lafferty,
\newblock ``Nonparametric regression and classification with joint sparsity
  constraints,''
\newblock in {\em NIPS}, 2009, pp. 969--976.

\bibitem{davis1997adaptive}
G~Davis, S~Mallat, and M~Avellaneda,
\newblock ``Adaptive greedy approximations,''
\newblock {\em Constructive approximation}, vol. 13, no. 1, pp. 57--98, 1997.

\bibitem{ekanadham2011sparse}
C~Ekanadham, D~Tranchina, and E~Simoncelli,
\newblock ``Sparse decomposition of transformation-invariant signals with
  continuous basis pursuit,''
\newblock in {\em IEEE ICASSP}, 2011, pp. 4060--4063.

\bibitem{fawzi2013classification}
A~Fawzi and P~Frossard,
\newblock ``Classification of unions of subspaces with sparse
  representations,''
\newblock in {\em Asilomar Conference on Signals, Systems and Computers}, 2013,
  pp. 1368--1372.

\bibitem{tropp2004greed}
J~Tropp,
\newblock ``Greed is good: Algorithmic results for sparse approximation,''
\newblock {\em IEEE Trans. on Inf. Theory}, vol. 50, no. 10, pp. 2231--2242,
  2004.

\bibitem{davenport2010analysis}
M~Davenport and M~Wakin,
\newblock ``Analysis of orthogonal matching pursuit using the restricted
  isometry property,''
\newblock {\em IEEE Trans. on Inf. Theory}, vol. 56, no. 9, pp. 4395--4401,
  2010.

\bibitem{peotta2007matching}
L~Peotta and P~Vandergheynst,
\newblock ``Matching pursuit with block incoherent dictionaries,''
\newblock {\em IEEE Trans. on Sig. Proc.}, vol. 55, no. 9, pp. 4549--4557,
  2007.

\bibitem{eldar2010block}
Y~Eldar, P~Kuppinger, and H~Bolcskei,
\newblock ``Block-sparse signals: Uncertainty relations and efficient
  recovery,''
\newblock {\em IEEE Trans. on Sig. Proc.}, vol. 58, no. 6, pp. 3042--3054,
  2010.

\bibitem{ba2012adaptive}
A~Ba, M~Sinn, Y~Goude, and P~Pompey,
\newblock ``Adaptive learning of smoothing functions: application to
  electricity load forecasting,''
\newblock in {\em NIPS}, 2012, pp. 2510--2518.

\bibitem{fan2012short}
S~Fan and RJ~Hyndman,
\newblock ``Short-term load forecasting based on a semi-parametric additive
  model,''
\newblock {\em IEEE Trans. on Power Systems}, vol. 27, no. 1, pp. 134--141,
  2012.

\bibitem{drucker1997support}
H~Drucker, C~Burges, L~Kaufman, A~Smola, V~Vapnik, et~al.,
\newblock ``Support vector regression machines,''
\newblock {\em NIPS}, vol. 9, pp. 155--161, 1997.

\bibitem{fan2008liblinear}
RE~Fan, KW~Chang, CJ~Hsieh, XR~Wang, and CJ~Lin,
\newblock ``Liblinear: A library for large linear classification,''
\newblock {\em The Journal of Machine Learning Research}, vol. 9, pp.
  1871--1874, 2008.

\bibitem{CC01a}
CC~Chang and CJ~Lin,
\newblock ``{LIBSVM}: A library for support vector machines,''
\newblock {\em \textit{ACM Transactions on Intelligent Systems and
  Technology}}, vol. 2, pp. 27:1--27:27, 2011.

\bibitem{cer_dataset_alternative}
``Smart metering information paper 4: Results of electricity cost-benefit
  analysis, customer behavior trials and technology trials,'' The Commission
  for Energy Regulation (CER), 2011.

\bibitem{fay200324}
D~Fay, J~Ringwood, M~Condon, and M~Kelly,
\newblock ``24-h electrical load data-a sequential or partitioned time
  series?,''
\newblock {\em Neurocomputing}, vol. 55, no. 3, pp. 469--498, 2003.

\bibitem{hong2014global}
T~Hong, P~Pinson, and S~Fan,
\newblock ``Global energy forecasting competition 2012,''
\newblock {\em International Journal of Forecasting}, vol. 30, no. 2, pp.
  357--363, 2014.

\end{thebibliography}
\end{document}